%% file: neurips_2022.tex
\documentclass{article}

\usepackage[nonatbib,preprint]{neurips_2022}

\usepackage[utf8]{inputenc} 
\usepackage[T1]{fontenc}    
\usepackage{hyperref}       
\usepackage{url}            
\usepackage{booktabs}       
\usepackage{amsfonts}       
\usepackage{nicefrac}       
\usepackage{microtype}      
\usepackage{xcolor}         
\usepackage{graphicx}
\usepackage{caption}
\usepackage{subcaption}

\usepackage{amsmath}
\usepackage{float}

\title{Rarity Score : A New Metric to Evaluate the Uncommonness of Synthesized Images}

\author{%
  Jiyeon Han \\
  Kim Jaechul Graduate School of AI \\
  KAIST\\
  \texttt{j.han@kaist.ac.kr} \\
   \And
   Hwanil Choi \\
   Kim Jaechul Graduate School of AI  \\
   KAIST\\
   \texttt{hwanil.choi@kaist.ac.kr} \\
  \AND
  Yunjey Choi  \\
  NAVER AI Lab \\
  \texttt{yunjey.choi@navercorp.com}\\
  \And
  Junho Kim \\
  NAVER AI Lab\\
  \texttt{jhkim.ai@navercorp.com}\\
  \And
  Jung-Woo Ha \\
  NAVER AI Lab\\
  \texttt{jungwoo.ha@navercorp.com}
  \And
  Jaesik Choi \\
  Kim Jaechul Graduate School of AI \\
  KAIST\\
  \texttt{jaesik.choi@kaist.ac.kr} \\
}
\usepackage{selectp}

\begin{document}

\maketitle

\begin{abstract}
 Evaluation metrics in image synthesis play a key role to measure performances of generative models. However, most metrics mainly focus on image fidelity. Existing diversity metrics are derived by comparing distributions, and thus they cannot quantify the diversity or rarity degree of each generated image. In this work, we propose a new evaluation metric, called `rarity score', to measure the individual rarity of each image synthesized by generative models. We first show empirical observation that common samples are close to each other and rare samples are far from each other in nearest-neighbor distances of feature space. We then use our metric to demonstrate that the extent to which different generative models produce rare images can be effectively compared. We also propose a method to compare rarities between datasets that share the same concept such as CelebA-HQ and FFHQ. Finally, we analyze the use of metrics in different designs of feature spaces to better understand the relationship between feature spaces and  resulting sparse images. Code will be publicly available online for the research community.

\end{abstract}

\input{text/intro}

\begin{figure}
  \centering
  
         \includegraphics[width=\textwidth]{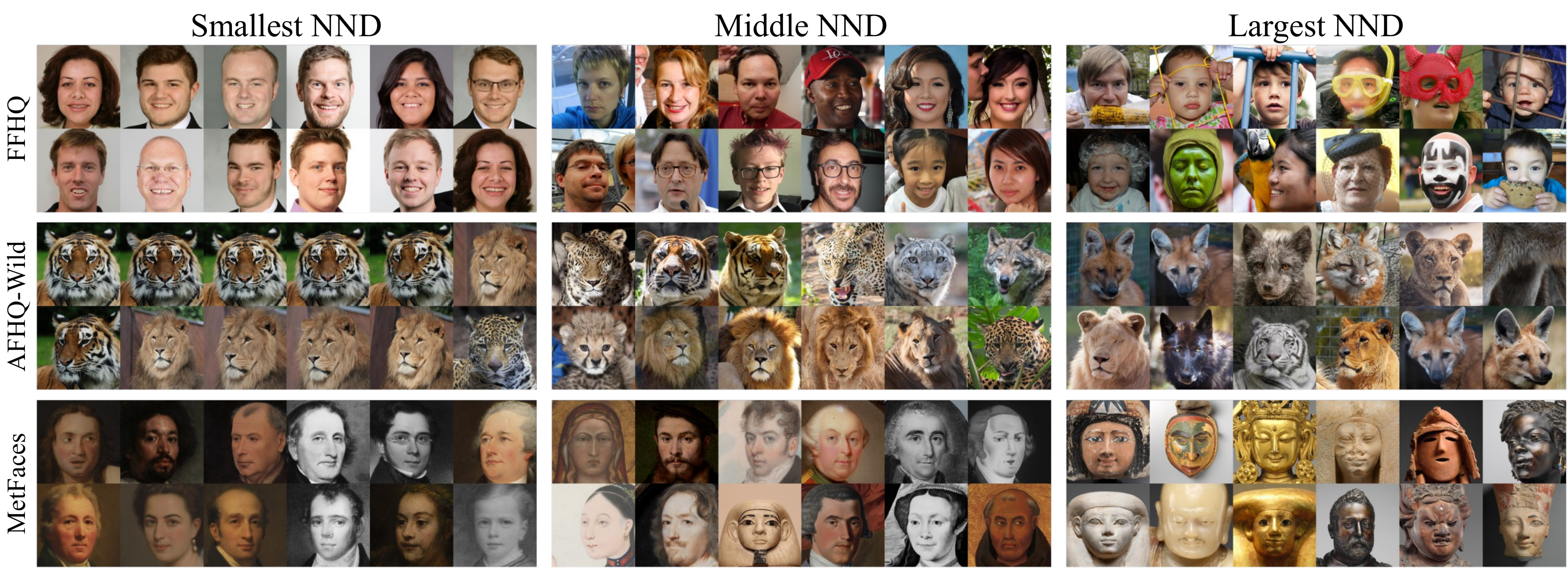}
         \caption{Real samples with the smallest nearest neighborhood distances (NNDs), middle NNDs, and the largest NNDs, respectively. For `Middle NND' column, the images are randomly selected among the middle-ranked 200 images. 
         }
         \label{fig:nnd}
  
\end{figure}

\section{Preliminaries}\label{sec:preliminaries}
\paragraph{Precision and Recall} 
Precision and recall are commonly used performance metrics in many areas including classification tasks or natural language processing. In specific, to quantify the performance of the generative models, precision measures the fraction of the fake distribution that can be generated from the real distribution. On the other hand, recall measures the fraction of the true distribution which can be reproduced by the fake distribution. Practically, $k$-nearest neighbor ($k$-NN) based method is proposed to estimate the real and fake manifolds \cite{kynkaanniemi2019improved}.       
For real samples $X_r\sim P_r$ and fake samples $X_g \sim P_g$, we first embed them in the high-dimensional feature space using pretrained DNNs such as VGG16 \cite{simonyanZ14a} or the image encoder of CLIP \cite{radford2021learning} to get sets of feature vectors $\mathbf{\Phi_r}$ and $\mathbf{\Phi_g}$, respectively. The real and fake manifolds are estimated by the sets of $k$-NN spheres of each sample as follows. 
\begin{equation}
\text{manifold}_k(\mathbf{\Phi}) = \bigcup_{\phi_i \in \mathbf{\Phi}} B_k(\phi_i,\mathbf{\Phi}), \quad B_k(\phi_i, \mathbf{\Phi})=\{\phi| d(\phi_i,\phi) \le NN_k(\phi_i,\mathbf{\Phi})\} 
\end{equation}
Here, $NN_k(\phi_i,\mathbf{\Phi})$ represents the distance between $\phi_i$ and its $k$-th nearest neighbor in $\mathbf{\Phi}$. $B_k(\phi_i,\mathbf{\Phi})$ is the $k$-NN sphere of $\phi_i$ with the radius of $NN_k(\phi_i,\mathbf{\Phi})$ defined as a set of all $\phi$ whose distance to $\phi_i$ is smaller than or equal to $NN_k(\phi_i,\mathbf{\Phi})$. For the distance metric $d$, we use L2 distance for the rest of the paper. Then, precision and recall are respectively defined as 
\begin{align}
\text{precision}(\mathbf{\Phi_r}, \mathbf{\Phi_g}) = \frac{1}{|\mathbf{\Phi_g}|}\sum_{\phi_j\in\mathbf{\Phi_g}}\mathbb{I}(\phi_j\in\text{manifold}_k(\mathbf{\Phi_r})),\\
\text{recall}(\mathbf{\Phi_r}, \mathbf{\Phi_g}) = \frac{1}{|\mathbf{\Phi_r}|}\sum_{\phi_i\in\mathbf{\Phi_r}}\mathbb{I}(\phi_i\in\text{manifold}_k(\mathbf{\Phi_g}))
\end{align}
where $\mathbb{I}$ is indicator function. Precision often maps with the fidelity and recall maps to the diversity of the generations. However, it is nontrivial to apply both precision and recall to an individual generation as they are designed to measure on the sets of samples.
To quantify the image quality of an individual generation, realism score is proposed. realism score measures the maximum of the inverse relative distance of a fake sample in a real $k$-NN sphere.  
\begin{equation}
\text{realism score}(\phi_j) = \max_{\phi_i \in \mathbf{\Phi_r}} \frac{NN_k(\phi_i,\mathbf{\Phi_r})}{d(\phi_i,\phi_j)}
\end{equation}
realism score is high when the distance between the given fake sample and a real sample is relatively small compared to the radius of the $k$-NN sphere of the real sample.
However, the realism score is not suitable to find rare samples as realism score can be large even if the radius of the $k$-NN sphere is small when the distance between fake sample and a real sample is also very small.  

\paragraph{Density and Coverage}
While precision and recall are commonly used, there are still several drawbacks that they are vulnerable to the outliers and computationally inefficient. To overcome the drawbacks, density and coverage are proposed \cite{naeem2020reliable}. While precision only counts whether the fake sample is inside the real manifold or not, density counts the number of the real $k$-NN spheres that contains the fake sample. 
\begin{equation}
    \text{density}(\mathbf{\Phi_r}, \mathbf{\Phi_g}) = \frac{1}{k|\mathbf{\Phi_g}|}\sum_{\phi_j\in\mathbf{\Phi_g}}\sum_{\phi_i\in\mathbf{\Phi_r}}\mathbb{I}(\phi_j\in B_k(\phi_i,\mathbf{\Phi_r}))
\end{equation}
If a fake sample is included in the multiple real $k$-NN spheres, it is more certain that the fake sample is inside the real manifold. Thus, density can be more robust to the outliers compared to the precision. On the other hand, similar to recall, coverage is relevant to the diversity of the generations. Coverage computes the number of real samples containing at least one fake sample in its $k$-NN sphere. 
\begin{equation}
    \text{coverage}(\mathbf{\Phi_r}, \mathbf{\Phi_g}) = \frac{1}{|\mathbf{\Phi_r}|}\sum_{\phi_i\in\mathbf{\Phi_r}}\mathbb{I}(\exists \phi_j\in \mathbf{\Phi_g}, \text{s.t., } \phi_j \in B_k(\phi_i, \mathbf{\Phi_r}))
\end{equation}
If a fake sample is sparsely located, the fake manifold can be exaggerated and the recall can be overestimated. As real manifold is known to have less outliers than the fake manifold, coverage can prevent such overestimation. Furthermore, coverage is more computationally efficient compared to the recall as coverage does not require the $k$-NN computations of fake samples.  
However, density and coverage are also not suitable to be applied on individual generations as they work on a set of generations. 

\paragraph{Truncation Trick}
The truncation trick is widely used in generative models to restrict latent sampling space to maintain the image quality \cite{kingma2018glow,brock2018large,karras2019style}. Typically, in StyleGAN based models \cite{karras2019style,karras2020analyzing,karras2020training}, the truncation trick is applied in the warped latent space $\mathcal{W}$ from the input latent space $\mathcal{Z}$ with $f:\mathcal{Z}\rightarrow \mathcal{W}$ as follows.
\begin{equation}
    \mathbf{w'} = \mathbf{\bar{w}}+\psi(\mathbf{w}-\mathbf{\bar{w}}), \quad \mathbf{\bar{w}}=\mathbb{E}_{z\in P(z)} [f(z)]
\end{equation}
When $\psi=1$, it is the same as not using the truncation trick. As $\psi$ decreases, the latent code shifts toward the mean latent code. When $\psi=0$, $\mathbf{w'}$ becomes $\mathbf{\bar{w}}$, which is the mean latent code in $\mathcal{W}$. 
Truncation trick is known to increase the fidelity at the expense of lowering the diversity of the generations.

\section{Rarity Score}\label{sec:method}

\begin{figure}[t]
  \centering
         \includegraphics[width=\textwidth]{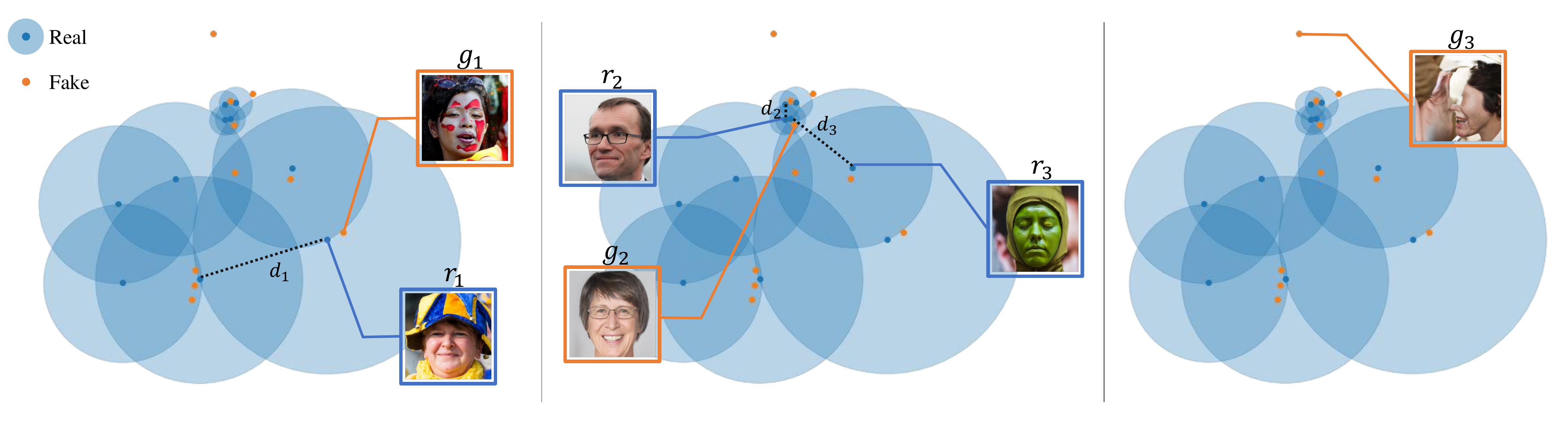}
         \makebox[0.35\linewidth][c]{\footnotesize{(a) Case 1: Single sphere}}\hfill
        \makebox[0.35\linewidth][c]{\footnotesize{(b) Case 2: Multiple spheres}}\hfill
        \makebox[0.29\linewidth][c]{\footnotesize{(c) Case 3: Out of real manifold}}\hfill
        \vspace{0.5mm}
         \caption{Real and fake images and their embeddings in the feature space when calculating the rarity score. The transparent circle represents the $k$-NN sphere of the corresponding real sample at the center. (a) When $g_1$ is only in $k$-NN sphere of $r_1$, $g_1$ is scored by the radius of $r_1$ ($d_1$). (b) When $g_2$ is in both $k$-NN spheres of $r_2$ and $r_3$, $g_2$ is scored by the smallest radius ($d_2$). Since $g_1$ has the higher score than $g_2$, $g_1$ looks more unique than $g_2$. (c) If a sample is out of real manifold as $g_3$, we cannot guarantee the fidelity of the image and thus exclude it from the scoring procedure.}
         \label{fig:main}
  
\end{figure}

To evaluate the rarity of individual generation, we propose a novel metric called \textit{rarity score}. Following the idea of \cite{kynkaanniemi2019improved,naeem2020reliable}, we use $k$-NN to represent the manifolds of real and generated samples as in Figure \ref{fig:main}. 

We hypothesize that ordinary samples would be closer to each other whereas unique and rare samples would be sparsely located in the feature space. 
Figure \ref{fig:nnd} supports our assumption. Figure \ref{fig:nnd} shows the real samples with both the smallest nearest neighbor distances (NNDs) and with the largest NNDs from the training dataset. For all datasets, samples with the smallest NNDs show representative and typical images. On the contrary, the samples with the largest NNDs have strong individuality and are significantly different from the typical images with the smallest NNDs. 
From this intuition, we propose the rarity score as follows,
\begin{equation}\label{eq:score}
    Rarity(\phi_g,\mathbf{\Phi_r}) = \min_{r, s.t. \phi_g\in B_k(\phi_r,\mathbf{\Phi_r}) }NN_k(\phi_r,\mathbf{\Phi_r}).
\end{equation}
  
  
  
  
\begin{figure}
  \centering
         \includegraphics[width=\textwidth]{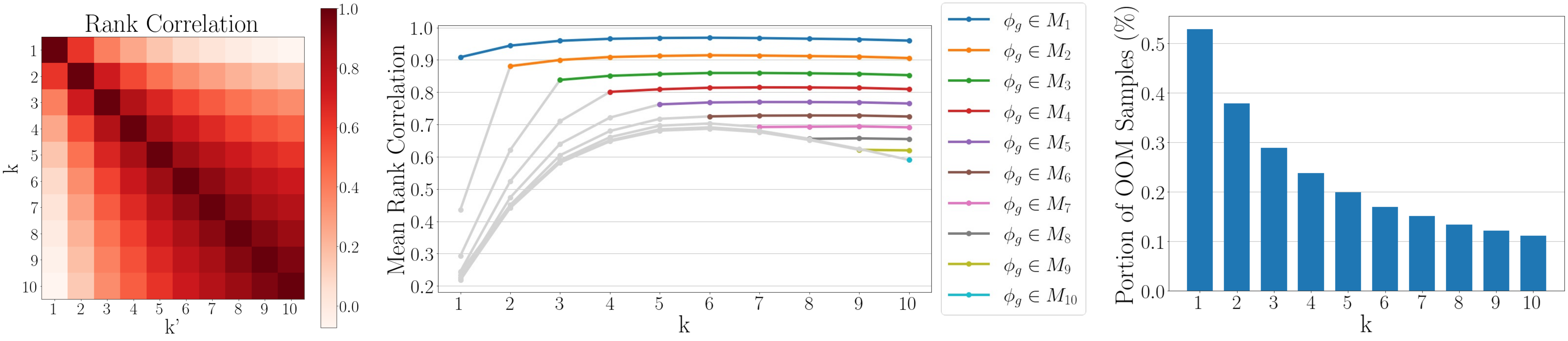}
         \caption{Rank correlations between $1\le k\le10$ for the rarity scores. Left : Rank correlation matrix between $1\le k\le10$ and $1\le k'\le10$ for the fake samples in $\{ \phi_g | \phi_g \in M_1$\} where $M_{i}$ stands for $\text{manifold}_i\mathbf{(\Phi_{r})})$ with $i$ as the $k$-NN parameter. Center : Mean rank correlation for all $k$s. Each plot represents the mean rank correlation for the fake samples in $\{ \phi_g | \phi_g \in M_{i}\}$. Right :  Portion of out of manifold (OOM) samples. }
         \label{fig:rank_cor}
  
\end{figure}

Equation (\ref{eq:score}) measures the rarity score of the given fake sample as the radius of $k$-NN sphere of a real sample which contains the given fake sample. In Case 1 in Figure \ref{fig:main}, $g_1$ is in the $k$-NN sphere of $r_1$, thus the rarity score of $g_1$ is the radius of the $k$-NN sphere of $r_1$, which is $d_1$. 
When there is more than one $k$-NN sphere containing the given fake sample, we use the minimum of the radii. If the radius of the smallest sphere is large, all other spheres containing the given fake sample will be at least as large as the smallest sphere. Then the given fake sample is at a low density region in the real manifold and we can think the given fake sample is extraordinary. By taking the minimum of the radii, we can prevent overestimating the sparsity of the given fake sample. For example, as in Case 2 of Figure \ref{fig:main}, $g_2$ is included in both $k$-NN sphere of $r_2$ and $r_3$. As the $k$-NN sphere of $r_3$ is large enough, it can cover $g_2$. However, the image of $g_2$ is far closer to $r_2$ with the smaller sphere than to $r_3$. Thus, $d_1$ represents the rarity of $g_1$ better than $d_3$. 
As the fidelity of the samples outside of the real manifold is not guaranteed, our metric discards the generations which is not contained in the real manifold $\text{manifold}_k(\mathbf{\Phi_r})$. For example, the generation $g_3$ in Figure \ref{fig:main} is an artifact and it is out of the real manifold. Thus the rarity score for $g_3$ is undefined. We provide the examples of generations out of the real manifold in Appendix.

  

\paragraph{The Choice of $k$} We show how $k$-NN parameter $k$ affects the proposed metric in terms of the Spearman rank correlation. Figure \ref{fig:rank_cor} shows the rank correlations among $1\le k\le 10$. The left heatmap shows the rank correlation matrix for the samples included in $M_1$ where $M_i$ indicates $\text{manifold}_i(\mathbf{\Phi_r})$, i.e., the real manifold built with $i$-NN spheres. The second plot in Figure \ref{fig:rank_cor} shows the average rank correlation between $k=1$ and $1\le k' \le10$. Each color indicates the target generations in $M_i$. Please note that $M_i \subset M_j$ for $i\le j$ and the rank correlation changes dramatically for the cases in $M_j\setminus M_i$. For example, for $M_3$, the mean rank correlations have high values $\ge 0.8$ for all $k\ge3$. While there exists slight drop in the mean rank correlation as $i$ increases, due to the change in the out-of-manifold (OOM) samples, we can see that the mean rank correlation is consistent for $k\ge i$. In other words, the rank of the proposed metric does not change much for different choices of $k$, and we can say the proposed metric is robust to the choice of $k$. However, if we decrease $k$, the number of fake samples out of the real manifold increases as we can see in the right graph in Figure \ref{fig:rank_cor}. Considering $k\ge3$ keep OOM samples under 30\%, we set $k=3$ for the rest of this paper. 

\begin{figure}[t]
  \centering
  
         \includegraphics[width=\textwidth]{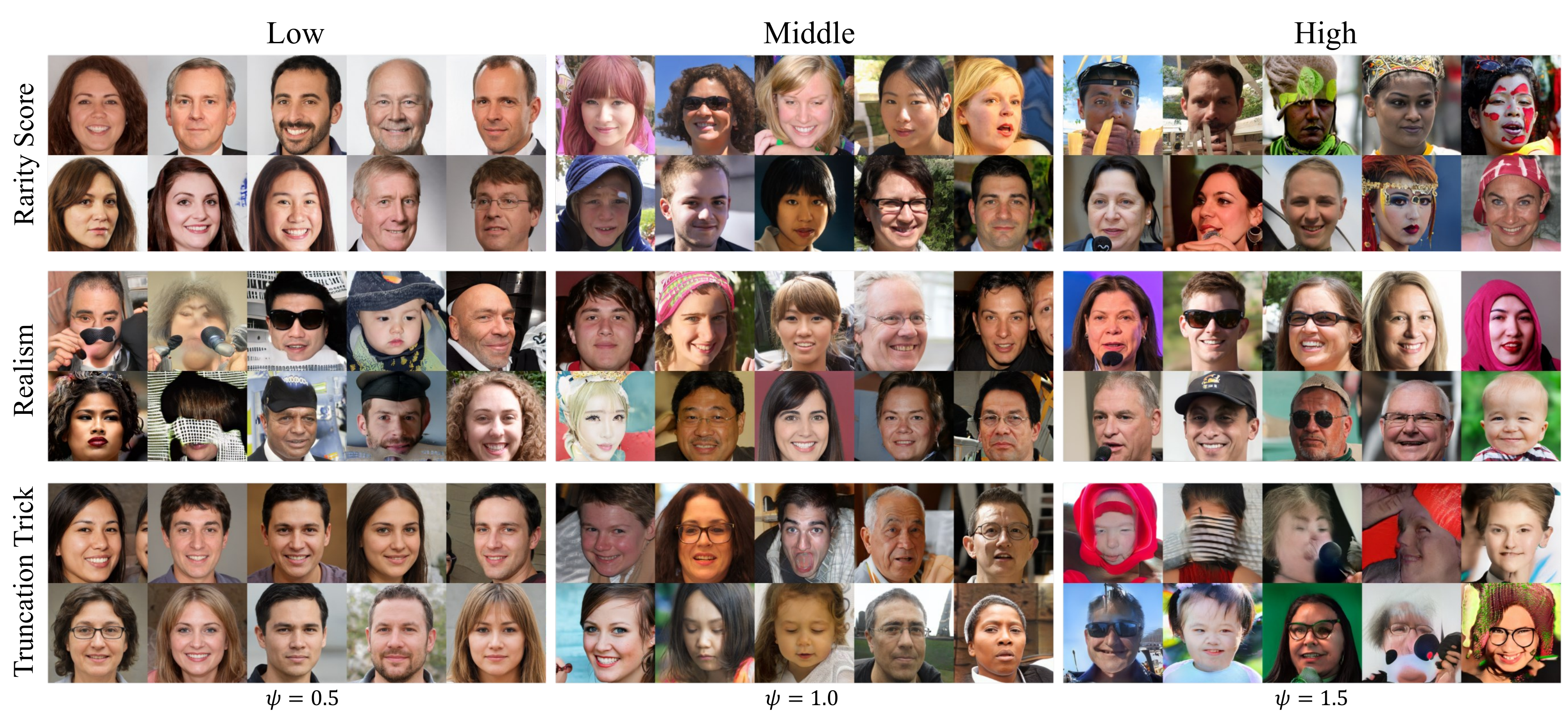}
         \caption{Images generated by StyleGAN2-FFHQ, sorted by the various methods. For the rarity score and the realism score, `Low' column presents the generations with the top 10 lowest scores, `High' column presents the generations with the top 10 highest scores, and `Middle' column presents the random generations in the middle of the top 10th and the bottom 10th. For the truncation trick, `Low', `Middle', `High' column show random generations with the truncation parameters $\psi=0.5, 1.0, 1.5$, respectively. }
         \label{fig:method_compar}
\end{figure}

\paragraph{Comparisons with Existing Methods}
Here, we compare our rarity with two existing methods that are most similar to ours: the realism score~\cite{kynkaanniemi2019improved} and the truncation trick~\cite{brock2018large}. The realism score is a metric which measures the fidelity of an individual generation. While the realism score takes into account the radius of the real $k$-NN sphere, the effect of the radius is depressed by the effect of the distance between the real and the fake sample. 
Realism score can be high for the samples with the high density region which is assumed to be typical and thus not suitable to measure rarity.
In Figure \ref{fig:method_compar}, while images with the high realism scores show high fidelity, the images with the high rarity scores show more extraordinary characteristics in the images. 

Truncation trick is proposed to maintain high fidelity of generations by filtering out the outskirt latent codes far from the mean latent code. This method can increase the fidelity of the images, but it can compromise its versatility.
If we use $\psi\ge 1$ to shift the latent code in the opposite direction from the mean latent code, we might get more diverse images. In this case, however, we cannot guarantee the fidelity of images as shown in Figure \ref{fig:method_compar}.  

\section{Experimental Results}
\label{sec:experiments}
In this section, we present the experimental results of the proposed metric for various standard GANs and datasets. 
We use $k=3$ for the manifold approximation which is used as a robust choice for the size of the neighborhood in \cite{kynkaanniemi2019improved}. We use 30k of real images to approximate the real manifold and calculate the rarity of 10k fake images. We use VGG16 as a feature extractor for all experiments except for Section 4.4. 
\subsection{Rarity of Individual Generation}
\label{subsec:rarity of individual generation}
\begin{figure}
  \centering
  
         \includegraphics[width=\textwidth]{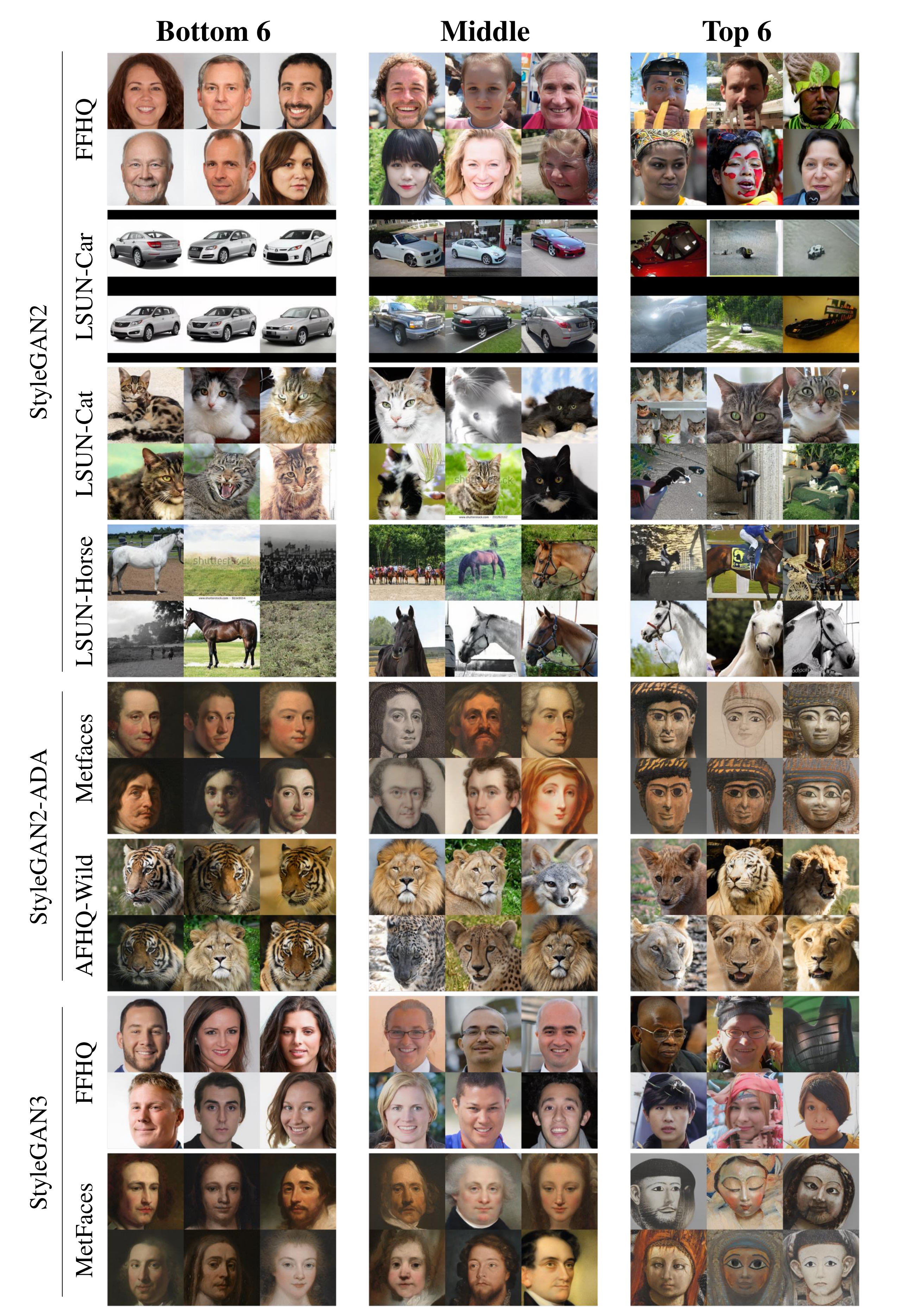}
         \caption{Bottom 6, Top 6, and the Middle generations of the proposed rarity score with various datasets.}
         \label{fig:metric1}
\end{figure}

Figure \ref{fig:metric1} shows generations with the top 6 highest, the top 6 lowest rarity scores and the generations in the middle. As expected, the generated images with the lowest rarity scores are the ones that are typical in the training dataset. 
On the other hand, the generations with the highest rarity scores are far from typical images and have various characteristics that are rare in the training dataset. 
For FFHQ dataset, for example, we can see clear and formal human face images in the `Bottom 6' column whereas we can see elaborated face images such as face with a colorful makeup or a fancy hat in the `Top 6' column. For AFHQ-Wild dataset, bottom images mostly show the faces of typical tigers while top images are mostly female lions with light-colored fur, which are far from the typical lion. Further, Metfaces dataset shows oil portraits in bottom images while top images show statues or flat paintings. However, for the datasets which are uncurated and not aligned such as LSUN datasets, the bottom 10 images are not clear as that of the aligned datasets because the training dataset contains a lot of noisy images for even human to accept.

\subsection{Comparisons between Generative Models}
Not only the proposed metric can be applied on an individual generation, it can be used to compare generative models on the same dataset which shares the real manifold. 
While there exist diversity metrics on the generative models such as LPIPS \cite{Zhang_2018_CVPR}, our method is different in that it focuses on the ability to generate unique and rare generations while the existing metrics focus on diversity between generations. 
\paragraph{Effects of Truncation Trick}
\begin{figure}
  \centering
  
         \includegraphics[width=\textwidth]{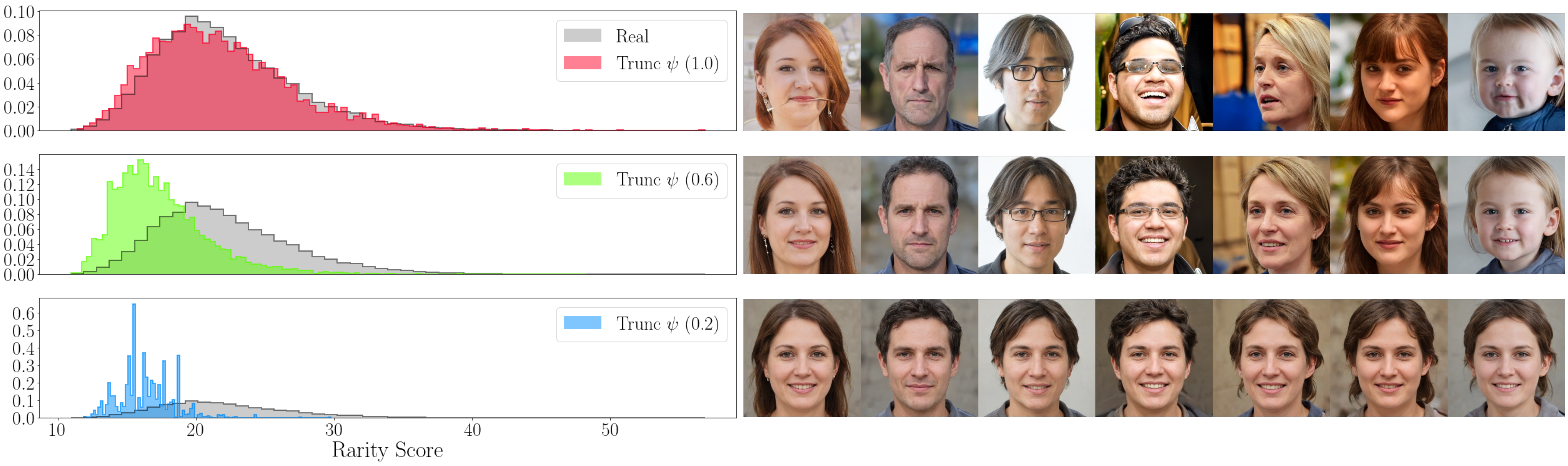}
         \caption{Histograms according to the truncation parameter $\psi$ and the corresponding generations for StyleGAN2-FFHQ. Left histograms show density of our rarity scores for each model with truncation parameter $\psi$. The smaller the $\psi$, the closer the $\psi$ is to the mean latent code. In this case, therefore, diversity of generated images is reduced. See the right images for the generations corresponding to each $\psi$. }
         \label{fig:histogram_psi}
\end{figure}

Before comparing the generative models, we first analyze how the distribution of the rarity scores changes with the truncation parameter. As truncation parameter is known to control between the diversity and the fidelity, it can be assumed that larger truncation parameters result in higher rarity scores. The histograms in Figure \ref{fig:histogram_psi} are consistent with the assumption. When the truncation parameter is small ($\psi\le 0.2)$, the images are almost same as the standard image generated from the mean latent code. Thus, the distribution of the rarity scores is concentrated on the specific values as many generations map to the same real $k$-NN sphere. When the truncation parameter gets larger, the generations become more diverse and the histogram of the rarity score shifts to the right. 

\paragraph{An Empirical Comparison between Generative Models}

\begin{figure}
  \centering
  
         \includegraphics[width=\textwidth]{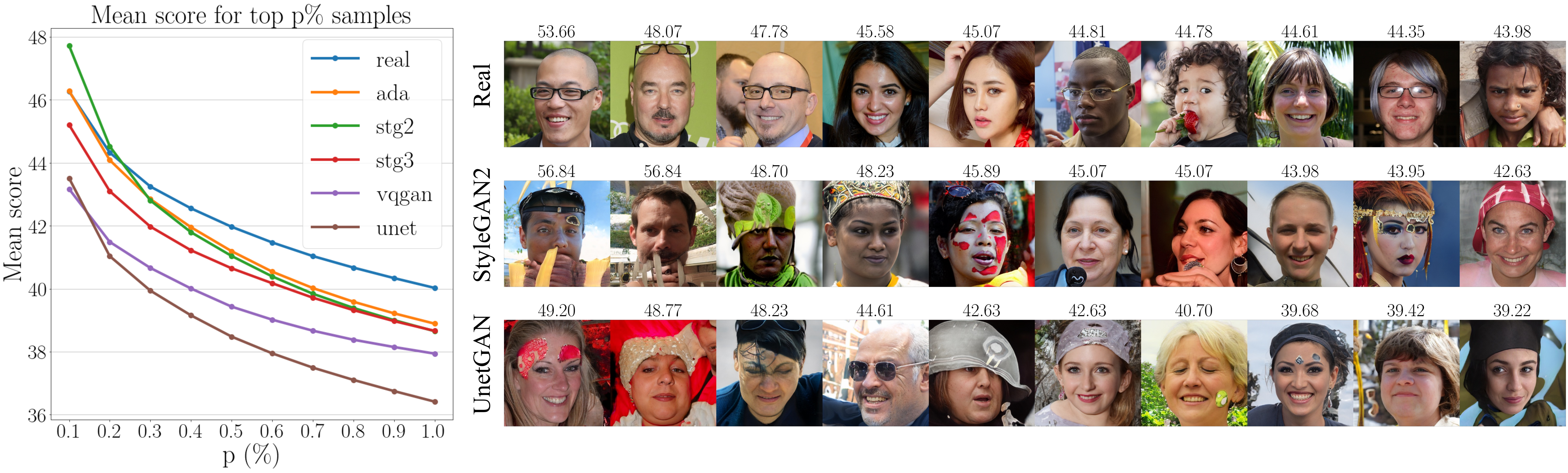}
         \caption{Mean scores of the top $p\%$ samples (left) and the top 10 images with the highest rarity scores (right). The number above each image represents the rarity score of the corresponding image. We investigate the mean scores of top $p\%$ samples for 5 models trained on FFHQ dataset; StyleGAN (stg), StyleGAN2 (stg2), StyleGAN3 (stg3), VQGAN (vqgan), and UnetGAN (unet).
         }
         \label{fig:model_compar}
  
\end{figure}

We further compare the generative models using the rarity score. Since our main focus is on the rare samples, we compare models by the mean rarity score of the top $p\%$ samples among 10000 samples. The left graph in Figure \ref{fig:model_compar} shows the mean scores of the various models as $p$ changes from 0.1 to 1.0. The images in the right column of Figure \ref{fig:model_compar} are the top 10 images with the highest scores, which corresponds to $p=0.1$. Under $p=0.1$, StyleGAN2 shows the highest mean score even when compared with the real images as we can see in the generations on the right. The images generated by StyleGAN2 are colorful and have diverse accessories, such as crown, mic, and hair band. On the other hand, the images from UnetGAN seem to have relatively less diversity. We can interpret that UnetGAN generates images more conservatively as its mean scores are lowest among the models for all $0.1\le p \le 1.0$.

\subsection{Comparisons between Similar Datasets}

\begin{figure}
  \centering
  
         \includegraphics[width=\textwidth]{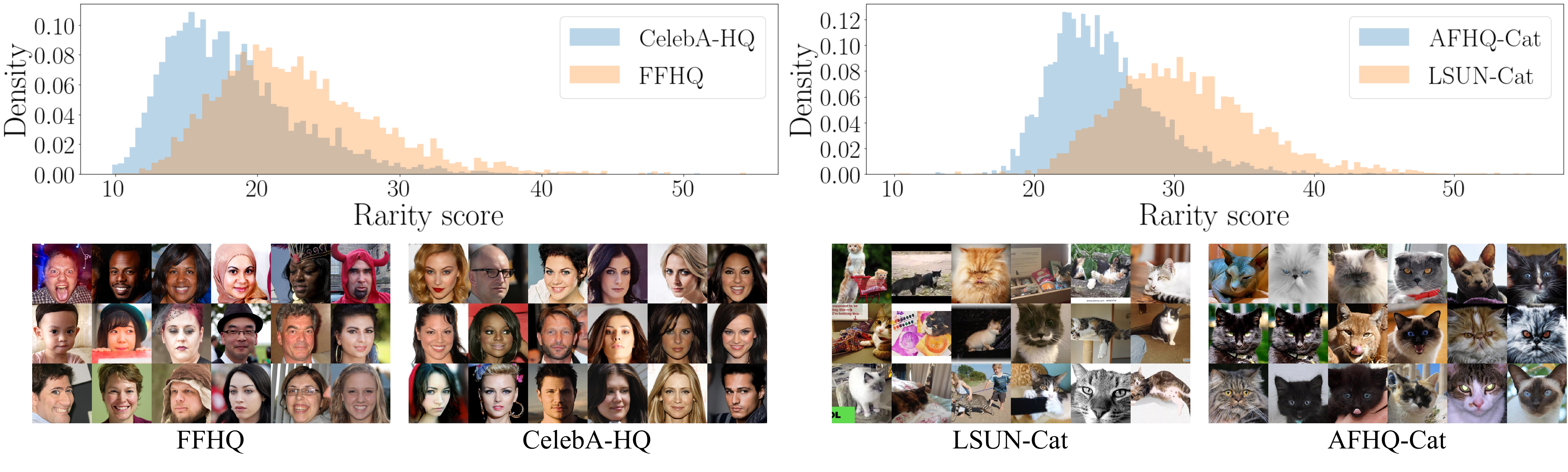}
         \caption{The density of rarity score for 10000 images sampled from datasets sharing the same concept (top) and images with the high rarity scores for each dataset (bottom). Rarity score is measured on the union of real manifolds for a pair of datasets. 
         }
         \label{fig:data_compare}
  
\end{figure}
We also propose a method to compare rarity between two different datasets which share the same concept, such as CelebA-HQ and FFHQ for human faces. Rarity scores of each dataset are calculated on union of two real datasets,  $\mathbf{\Phi_r}= \mathbf{\Phi_{r_1}} \cup \mathbf{\Phi_{r_2}}$. For example, when we calculate the scores of human face datasets such as CelebA-HQ and FFHQ, the real manifold $\text{manifold}_k(\mathbf{\Phi_r} )$ is $\text{manifold}_k(\mathbf{\Phi_{CelebA-HQ}} \cup \mathbf{\Phi_{FFHQ}})$. Figure \ref{fig:data_compare} shows comparisons of two sets of datasets which have the same concept on normalized rarity score. In the case of human face dataset, the density of CelebA-HQ is skewed left while the density of FFHQ is widely distributed to the right. It means that FFHQ has more uncommon images than CelebA-HQ. In other words, there are not many uncommon images in CelebA-HQ. This makes sense because CelebA-HQ is collected among the celebrities while FFHQ is collected among the public with the considerable variation in terms of age, ethnicity and image background. The images in Figure \ref{fig:data_compare} are top scoring images in each dataset. Since the maximum rarity score of FFHQ is greater than that of CelebA-HQ, the top images of FFHQ are relatively more uncommon. Similarly, LSUN-Cat and AFHQ-Cat are both datasets on cat images 
where LSUN-Cat is uncurated cat images and AFHQ-Cat is curated with close-up cat faces only. In Figure \ref{fig:data_compare}, we can confirm that LSUN-Cat is more diverse than AFHQ-Cat as the rarity score distribution of AFHQ-Cat is more left and concentrated compared to the distribution of AFHQ-Cat.


\subsection{A Study on Feature Extractors}
\begin{figure}
  \centering
  
         \includegraphics[width=\textwidth]{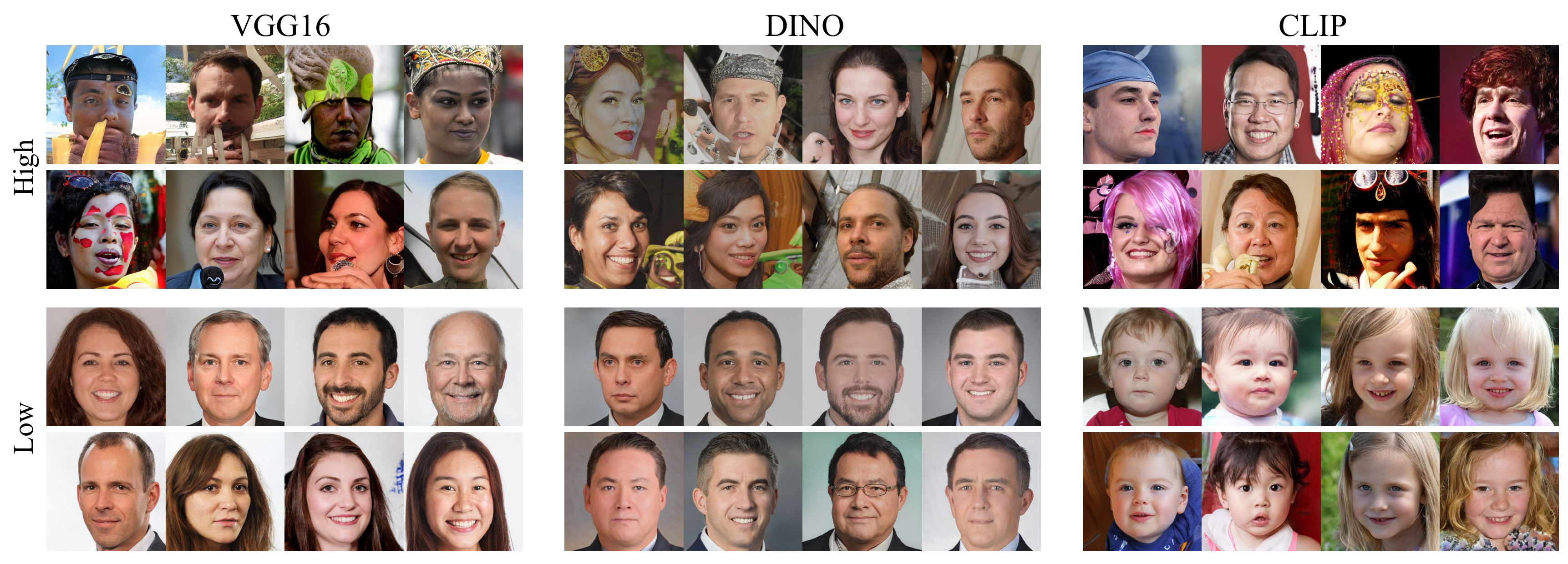}
         \caption{StyleGAN2 generations of FFHQ with the highest rarity scores and the lowest rarity scores for various feature extractors. 
         For the low score images, however, as CLIP is one of the self-supervised models, it captures semantics rather than structures or patterns unlike the other classifiers. 
         }
         \label{fig:backbone}
  
\end{figure}
Since the proposed metric operates in the feature space, the perspective of the rarity may vary depending on which feature extractor is used. Figure~\ref{fig:backbone} presents images with high and low rarity scores for diverse feature extractors to verify the effects of the extractors. As a classifier trained on the Imagenet dataset, VGG16 focuses more on the objects and the patterns. There are many images with the colorful face painting among the high score samples with VGG16. On the other hand, ViT models such as DINO~\cite{caron2021emerging} are known to focus more on the structure compared to the CNN-based models \cite{park2022how}. This can be seen from the second column of Figure \ref{fig:backbone}. The DINO-based high score images have various face angles while the low score images are mostly front-facing. In the third column of Figure \ref{fig:backbone}, the baby faces appear in the low score images unlike the other two models which have formal adult images in the low score. This can be explained from the previous study that the CLIP model~\cite{radford2021learning}, which is a self-supervised image-text multi-modal model, can better capture semantics compared to the classifiers. We conjecture that the model may consider the baby faces more similar between them than between the other images because baby faces are less distinctive and share a lot common properties compared to the adult faces.

\section{Conclusion}
In this paper, we propose a new evaluation metric called \textit{rarity score} to measure how rare a synthesized image is based on the real data distribution. Each generated image is scored by estimating the density around the target image on the real manifold. Our rarity score enables several following tasks that existing metrics, such as FID and Precision\&Recall, cannot do. First, the proposed metric can evaluate rarity or diversity of an individual image rather than a set of images. Second, the proposed metric can be applied to compare between generative models or datasets that share the same concept in terms of the density of the rare samples. In addition, we provide studies on feature extractors that the viewpoint of commonness for each feature extractor varies depending on its structure (e.g. fully CNN, ViT) or how it was learned (e.g. fully supervised, self-supervised). 

\begin{ack}
This work was supported by KAIST-NAVER Hypercreative AI Center, Institute of Information \& communications Technology Planning \& Evaluation (IITP) grant
funded by the Korea government (MSIT) (No.2017-0-
01779, XAI, No.2022-0-00184, Development and Study of AI Technologies to Inexpensively Conform to Evolving Policy on Ethics and No.2019-0-00075, Artificial Intelligence Graduate School Program (KAIST))
\end{ack}

\bibliography{neurips_2022}
\bibliographystyle{neurips_2022}

\appendix
\input{text/appendix}



\end{document}

%% file: text/intro.tex
\section{Introduction}
\label{sec:intro}
Generative Adversarial Networks (GANs) \cite{Goodfellow2014generative} have achieved significant advancement over the past several years, enabling many computer vision tasks such as image manipulation~\cite{kim2021exploiting, jahanian2019steerability, shen2020interfacegan, harkonen2020ganspace, bau2018gan}, domain translation~\cite{isola2017image, zhu2017cyclegan, choi2018stargan, choi2020stargan, Kim2020U-GAT-IT, kim2019tag2pix}, and image or video generation~\cite{kim2021feature, kim2022c3gan, karras2019style, karras2020analyzing, karras2020training, karras2021alias, skorokhodov2021stylegan, yu2022generating, tian2021good, tulyakov2018mocogan}. In addition, the resolution and quality of images synthesized by generative models have seen rapid improvement recently in terms of quantitative metrics~\cite{binkowski2018demystifying, heusel2017gans, salimans2016improved}. 

As standard evaluation metrics, inception score (IS)~\cite{salimans2016improved}, kernel inception distance (KID)~\cite{binkowski2018demystifying}, and Frech\'{e}t inception distance (FID)~\cite{heusel2017gans} are prevalent for evaluating the quality of images synthesized by generative models. 
These metrics evaluate the discrepancy between generated and real image sets on the feature space characterized by a generative model with respect to diversity and fidelity. The fidelity represents the quality of the generated image, and the diversity indicates that the generator creates images without mode collapse, similar to the distribution of the training datasets. To improve the fidelity and diversity, it is required that the distribution of generated images is similar to the real image distribution~\cite{kynkaanniemi2019improved, naeem2020reliable}. 

This learning scheme leads the generator not to synthesize much rare samples which are unique and have strong characteristics that do not account for a large proportion of the real image distribution. Examples of rare samples from public datasets include people with various accessories in FFHQ \cite{karras2019style}, white animals in AFHQ \cite{choi2020stargan}, and uncommon statues in Metfaces~\cite{karras2020training}. The ability to generate rare samples is important not only because it is related to the edge capability of the generative models, but also because uniqueness plays an important role in the creative applications such as virtual humans. However, the qualitative results of several recent studies seldom contain these rare examples. We conjecture that the nature of the adversarial learning scheme forces generated image distribution similar to that of a training dataset. Thus, images with clear individuality or rareness only take a small part in images synthesized by the models.
If we consider the distribution of training data, it is clear that low density areas are poorly represented and thus likely to be difficult for the generator to reproduce.
This is known as a significant open problem in most generative model learning\cite{karras2019style}. As a result, several recent works have focused on improving the average image quality by drawing latent vectors from a simplified sampling space although these tricks reduce the degree of variation. One of the techniques is the truncation trick proposed by BigGAN~\cite{brock2018large}, which reliably generates images normally distributed in the training dataset through a new latent vector with a truncation scale parameter $\psi$ between one latent vector and mean latent vectors. However, it is not intended to generate rare samples in training datasets, but rather to synthesize typical images more stably. We hypothesize that existing generative models will be able to produce samples richer in the real data distribution if the generator can be induced to effectively produce rare samples. 

In this paper, we propose a novel generative model evaluation metric that can represent the generative capabilities of rare samples in generative models as scores (a.k.a. rarity score). Our metric contributes to classifying rare images and typical images similar to those frequently observed in training datasets. The proposed rarity metric alleviates the open challenge problem of sparse generation of rare samples, and helps generative models synthesize rare samples without significantly losing fidelity. Additionally, we have conducted comparative experiments on which of the previous state-of-the-art models produce more rare samples with preserving quality performances. Our contributions can be summarized as follows:
\begin{itemize}
    \item We propose a novel metric to quantify the rarity of individual generation which is infeasible from the existing metrics. Using the proposed metric, generations with the desired degree of rarity can be sampled. 
    \item We show that the proposed metric can be used to compare between generative models in terms of the rare generation. The proposed metric can further be used to compare between datasets that share the same concept such as CelebA-HQ and FFHQ datasets.
    \item We show the the proposed metric can be applied on top of the various feature spaces with different point of views, by analyzing the sampled generations.
\end{itemize}


%% file: text/appendix.tex
\newpage
\section{Rank Correlations}
\label{app:rank}
\begin{figure}[H]
     \centering
     \begin{subfigure}[b]{0.3\textwidth}
         \centering
         \includegraphics[width=\textwidth]{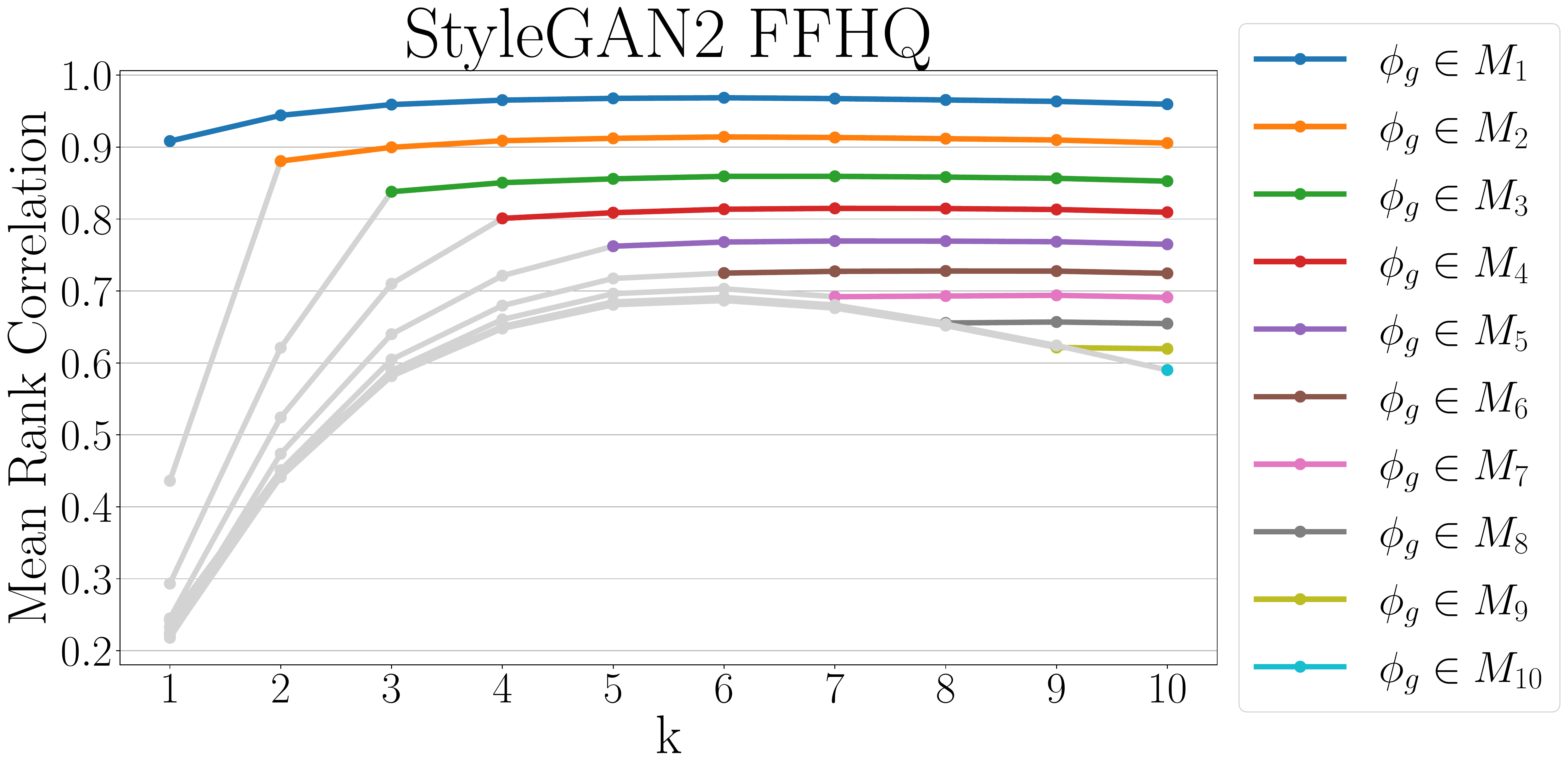}
         \caption{StyleGAN2 FFHQ}
         \label{fig_apdx:rank_ffhq}
     \end{subfigure}
     \hfill
     \begin{subfigure}[b]{0.3\textwidth}
         \centering
         \includegraphics[width=\textwidth]{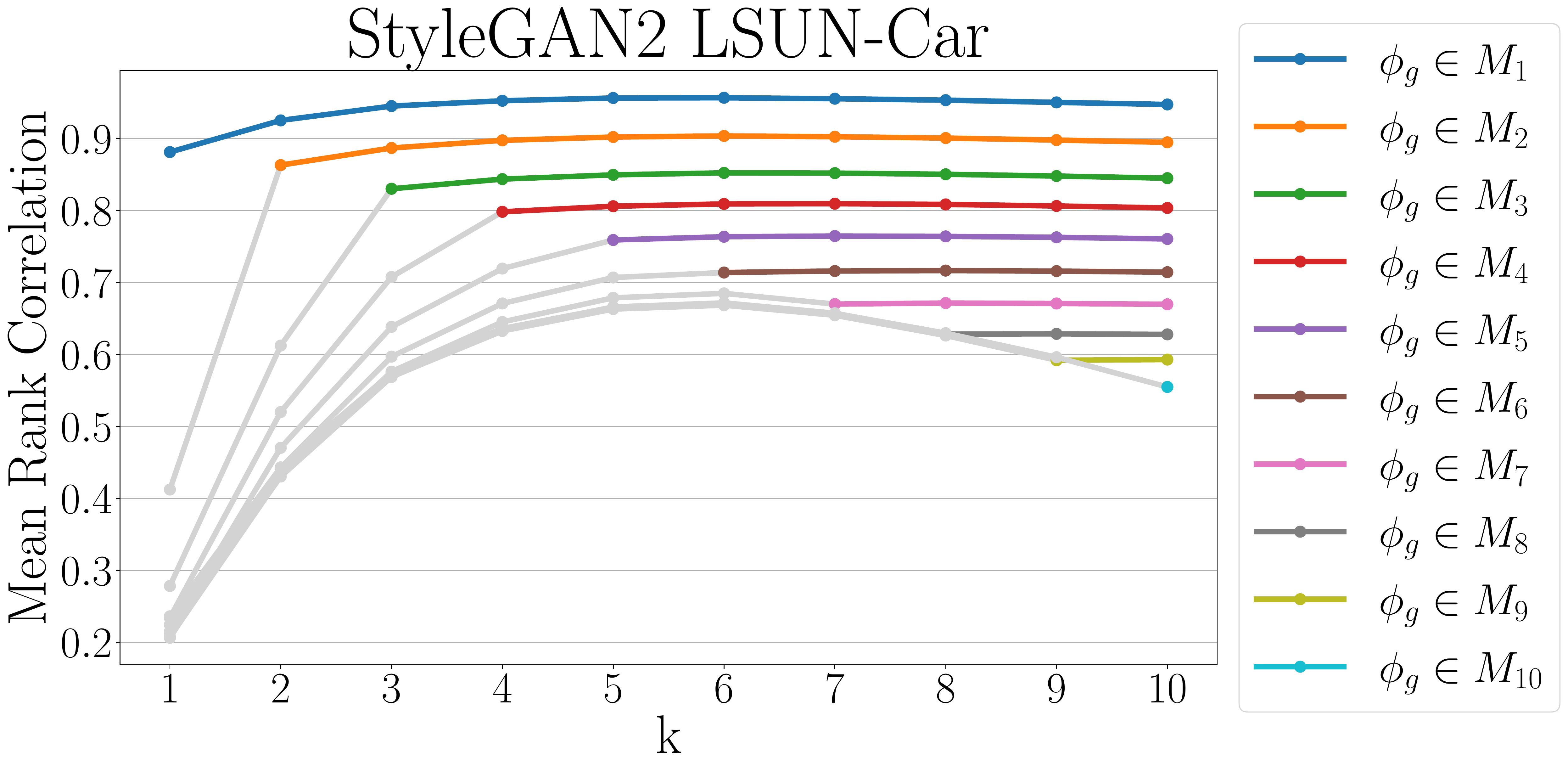}
         \caption{StyleGAN2 LSUN-Car}
         \label{fig_apdx:rank_car}
     \end{subfigure}
     \hfill
     \begin{subfigure}[b]{0.3\textwidth}
         \centering
         \includegraphics[width=\textwidth]{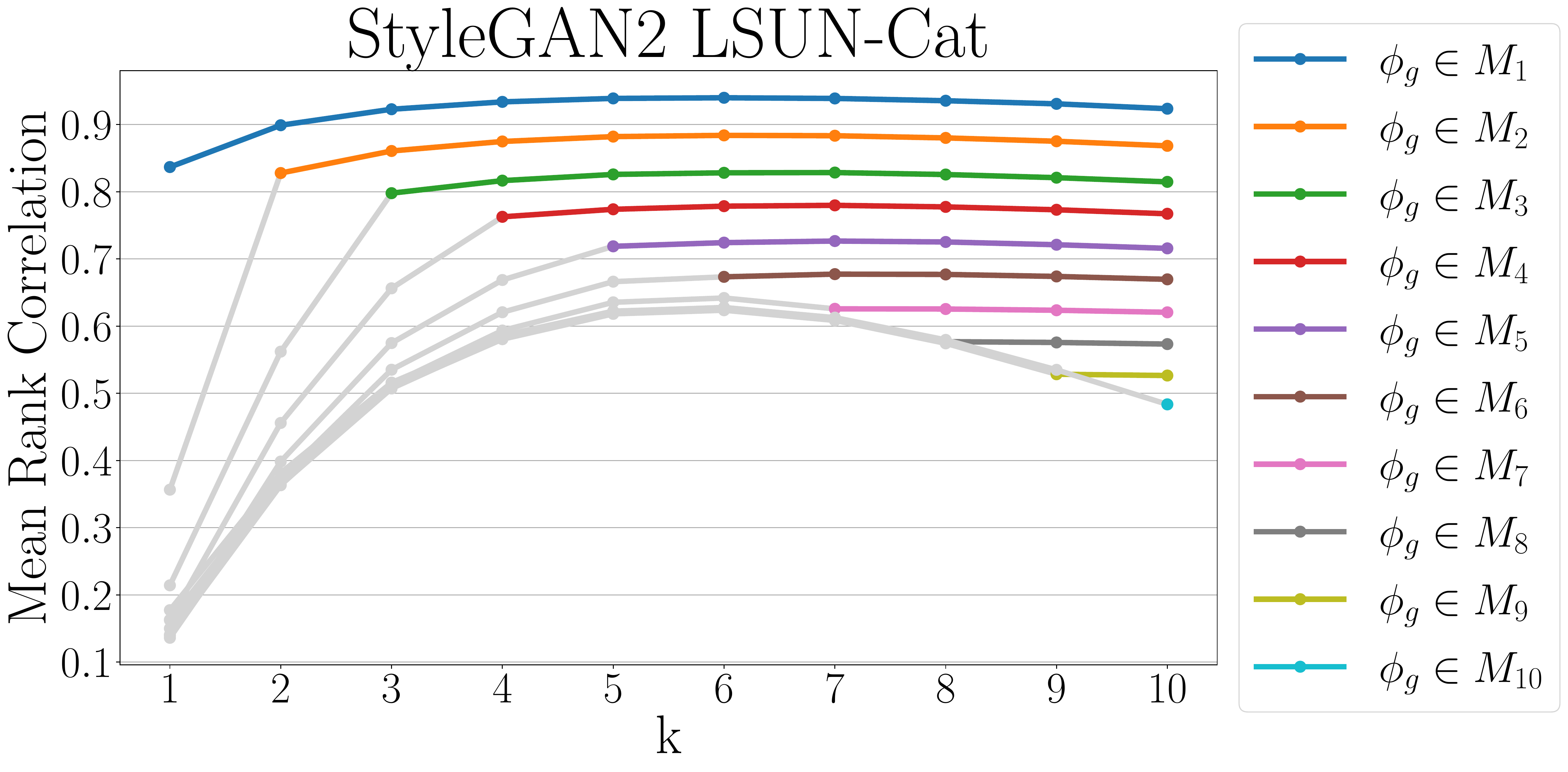}
         \caption{StyleGAN2 LSUN-Cat}
         \label{fig_apdx:rank_cat}
     \end{subfigure}
     \hfill
     \begin{subfigure}[b]{0.3\textwidth}
         \centering
         \includegraphics[width=\textwidth]{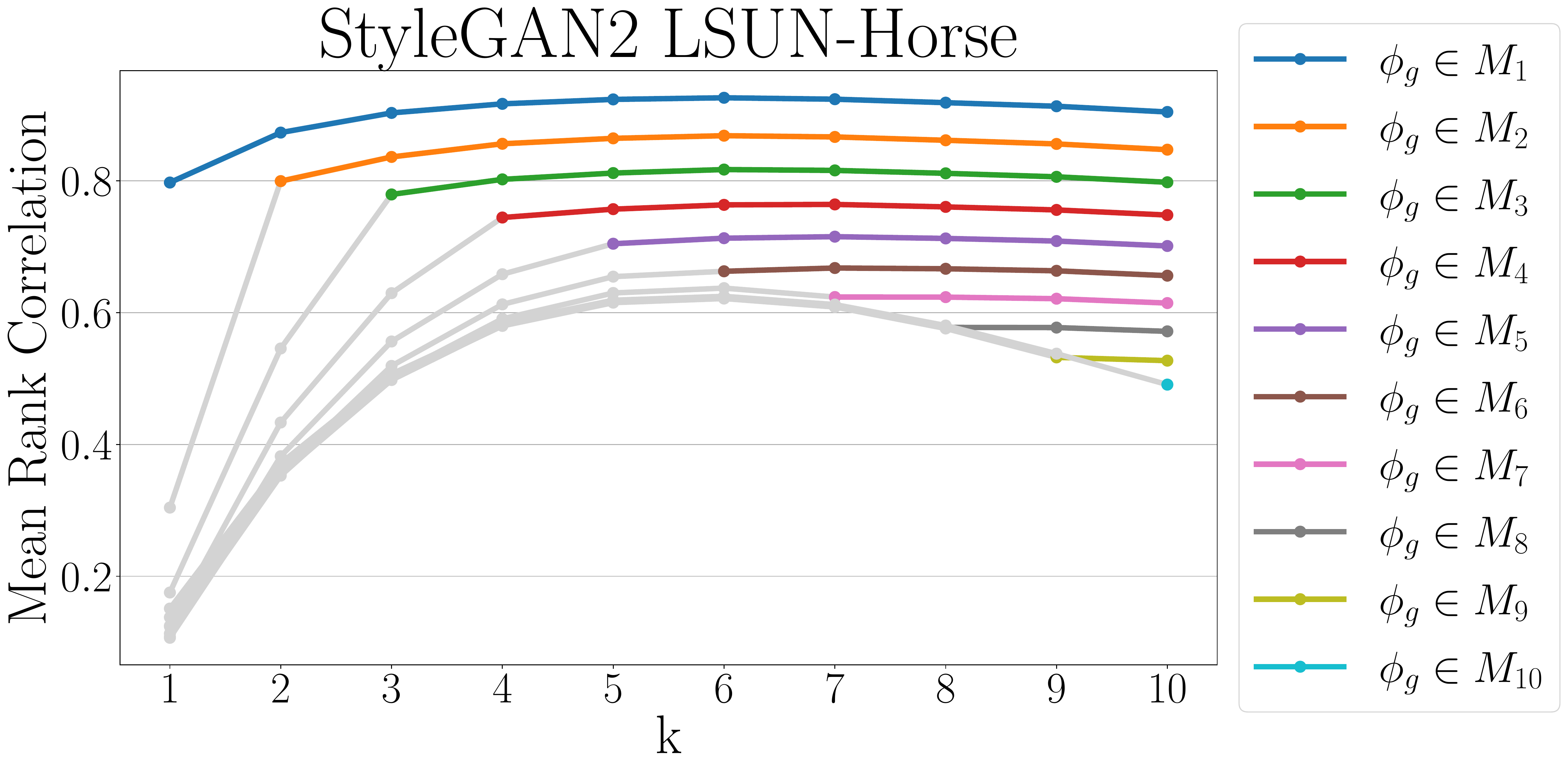}
         \caption{StyleGAN2 LSUN-Horse}
         \label{fig_apdx:rank_horse}
     \end{subfigure}
     \hfill
     \begin{subfigure}[b]{0.3\textwidth}
         \centering
         \includegraphics[width=\textwidth]{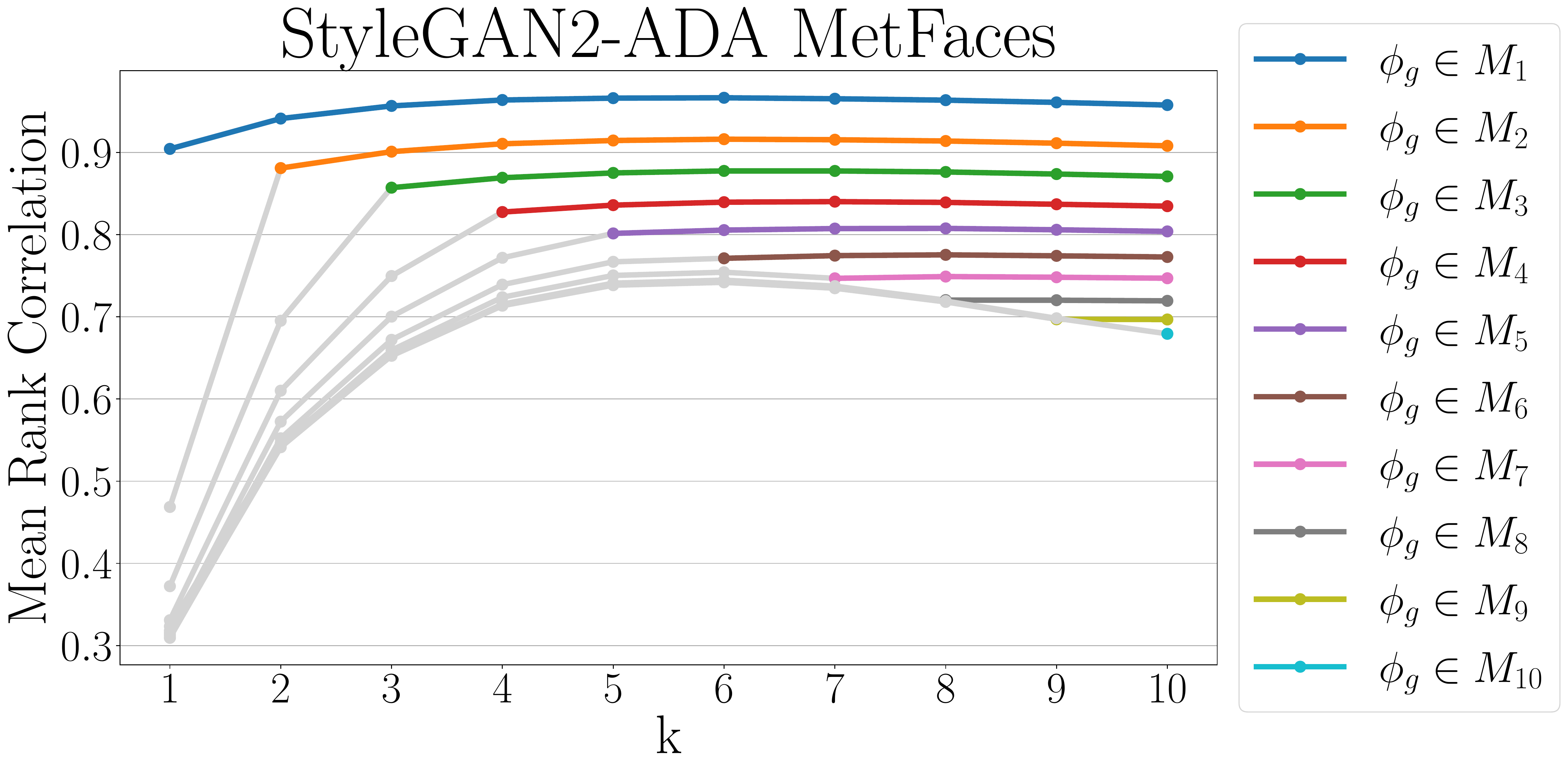}
         \caption{STG2-ADA MetFaces}
         \label{fig_apdx:rank_metface}
     \end{subfigure}
     \hfill
     \begin{subfigure}[b]{0.3\textwidth}
         \centering
         \includegraphics[width=\textwidth]{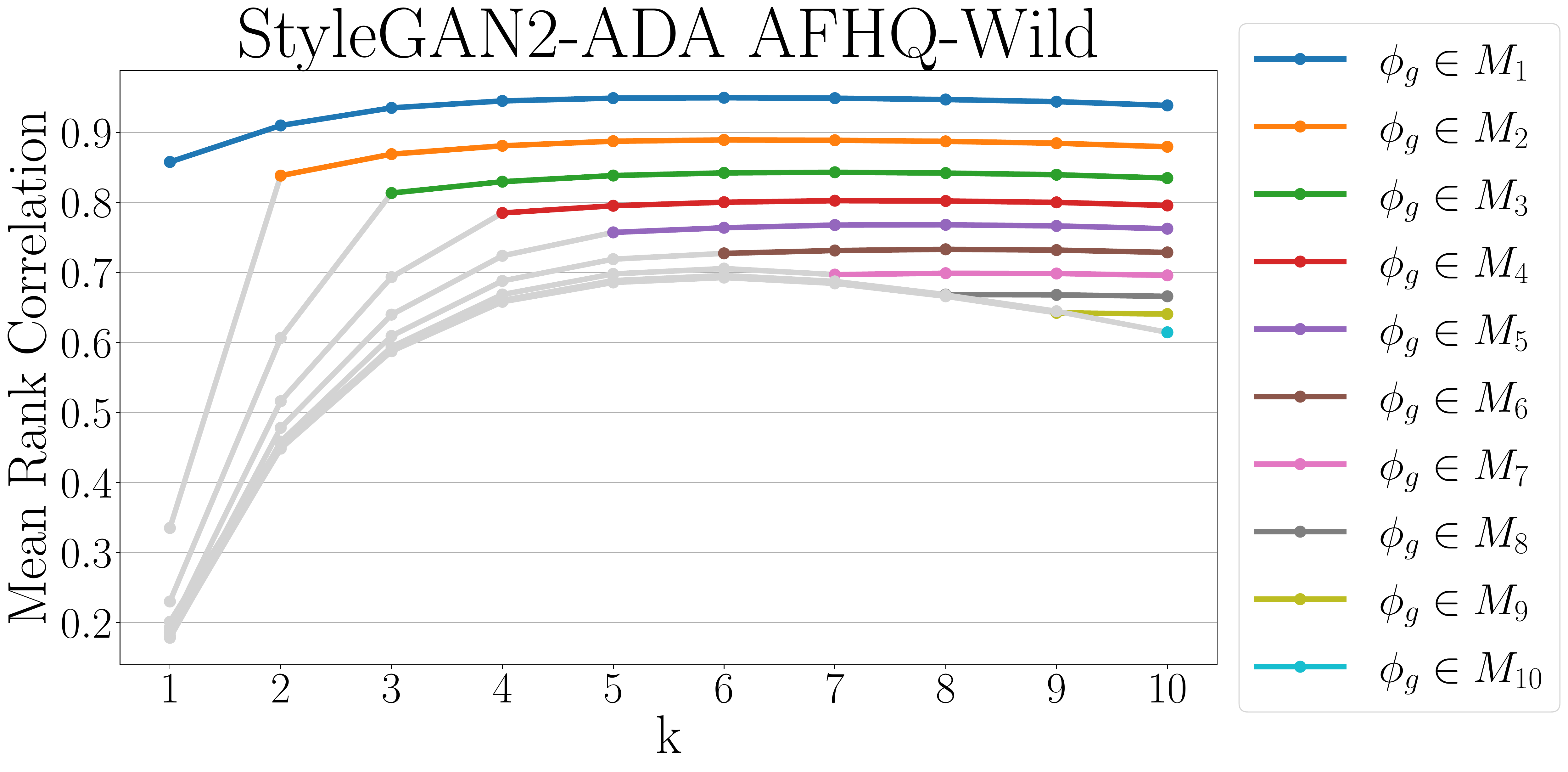}
         \caption{STG2-ADA AFHQ-Wild}
         \label{fig_apdx:rank_afhqwild}
     \end{subfigure}
        \caption{Rank correlations for the various models. STG2 stands for StyleGAN2.}
        \label{fig_adpx:rank}
\end{figure}

\section{Distributions According to the Truncation Parameter}
\begin{figure}[H]
  \centering

         \includegraphics[width=0.8\textwidth]{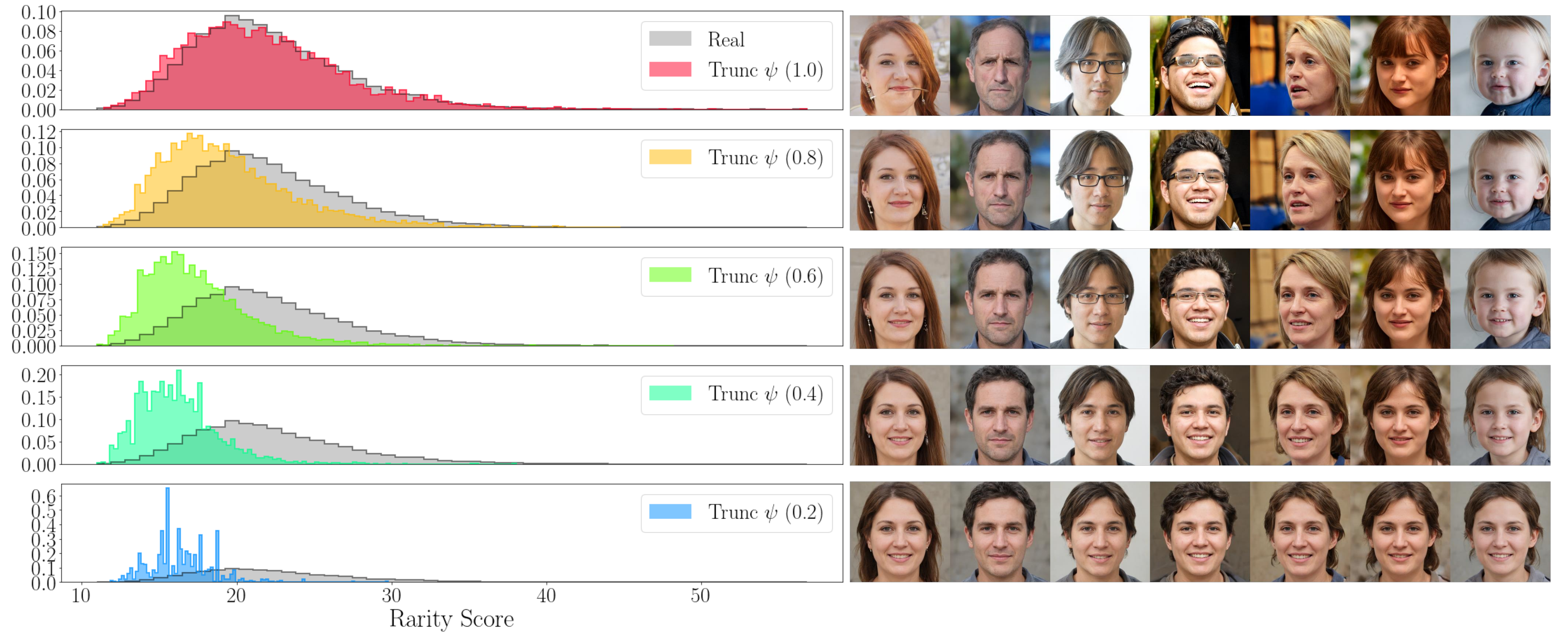}
         \caption{Histograms of the rarity score for 10000 generations varying the trunctaion parameter $\psi$ from 1.0 to 0.2. As expected, the distribution moves to the left as $\psi$ gets smaller (less diverse).
         }
         \label{fig_apdx:trunc}

\end{figure}    

\section{Comparisons between Models}
\begin{figure}[H]
  \centering
         \includegraphics[width=0.9\textwidth]{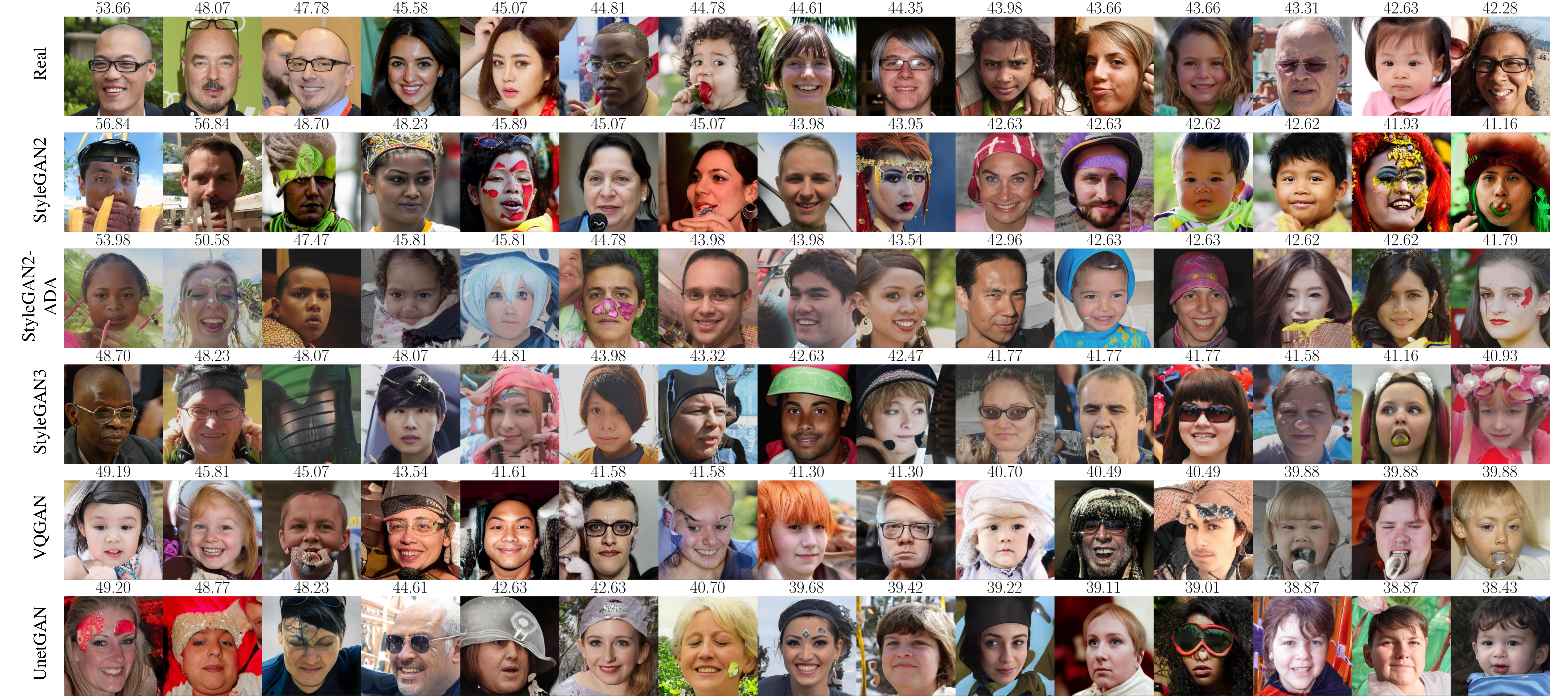}
         \caption{Top 15 samples of the highest rarity scores for the models trained on FFHQ dataset.}
         \label{fig_apdx:model_comparison}
\end{figure}    

\section{Real Samples Ordered by NND and Fake Samples Ordered by Rarity Score}
\label{app:qual}
\begin{figure}[H]
  \centering

         \includegraphics[width=0.95\textwidth]{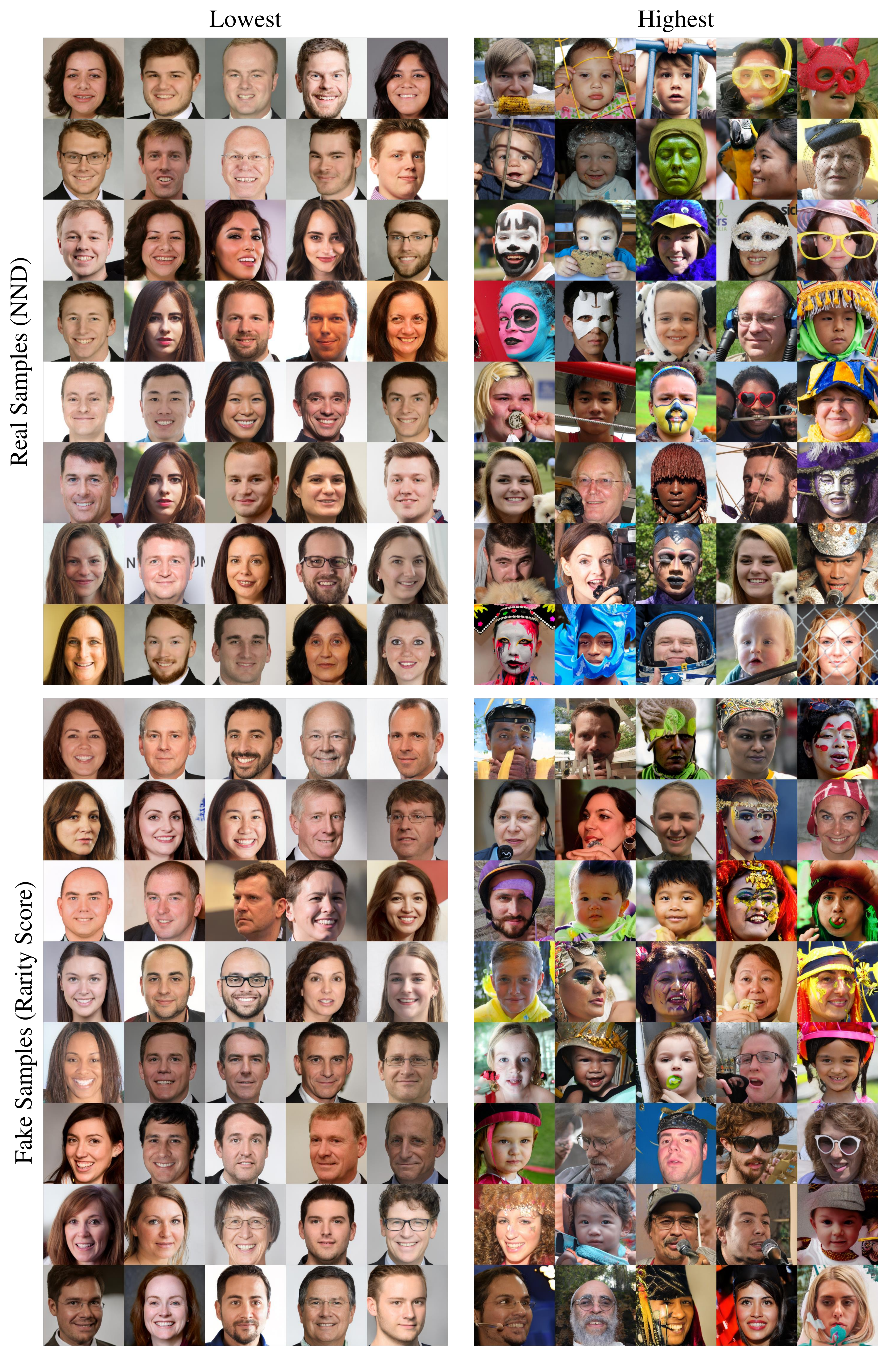}
         \caption{Real samples with the lowest/highest NND from FFHQ dataset (top) and fake samples generated from StyleGAN2 with the lowest/highest rarity score (bottom).
         }
         \label{fig_apdx:qual_ffhq}

\end{figure}    
  
\begin{figure}[H]
  \centering
  
         \includegraphics[width=\textwidth]{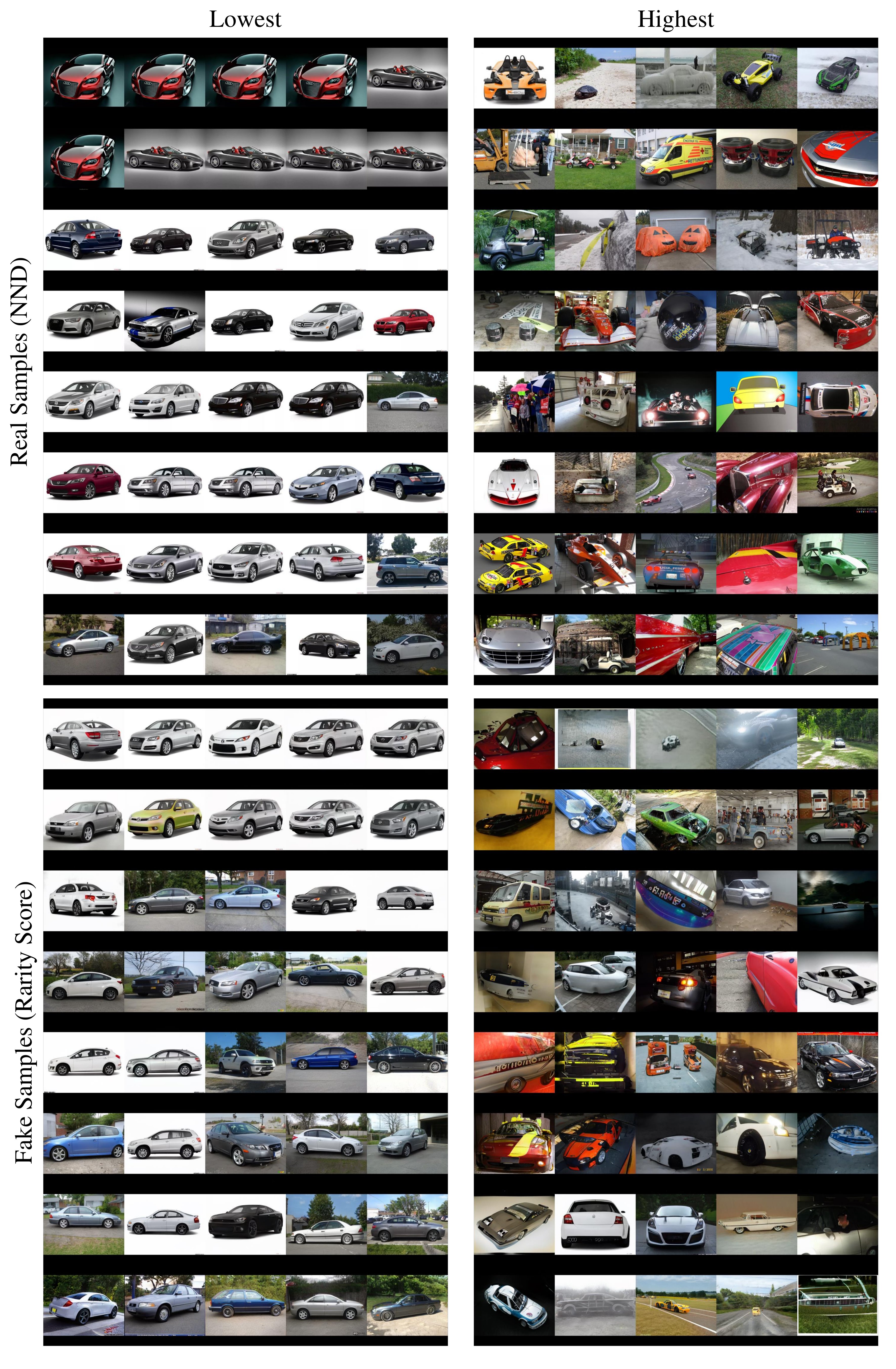}
         \caption{Real samples with the lowest/highest NND from LSUN-Car dataset (top) and fake samples generated from StyleGAN2 with the lowest/highest rarity score (bottom).
         }
         \label{fig_apdx:qual_car}
  
\end{figure}
\begin{figure}[H]
  \centering
  
         \includegraphics[width=\textwidth]{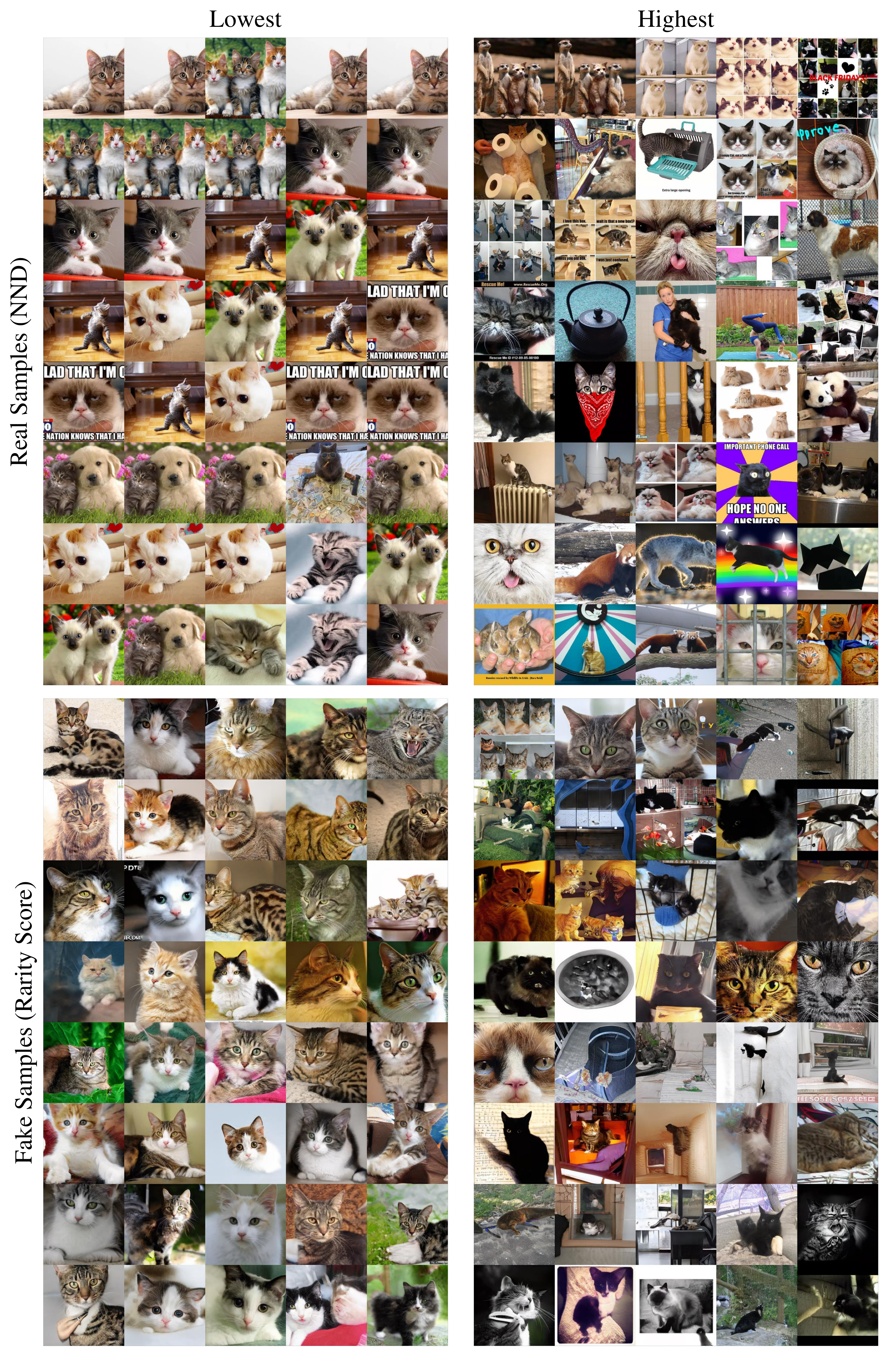}
         \caption{Real samples with the lowest/highest NND from LSUN-Cat dataset (top) and fake samples generated from StyleGAN2 with the lowest/highest rarity score (bottom).
         }
         \label{fig_apdx:qual_cat}
  
\end{figure}
\begin{figure}[H]
  \centering
  
         \includegraphics[width=\textwidth]{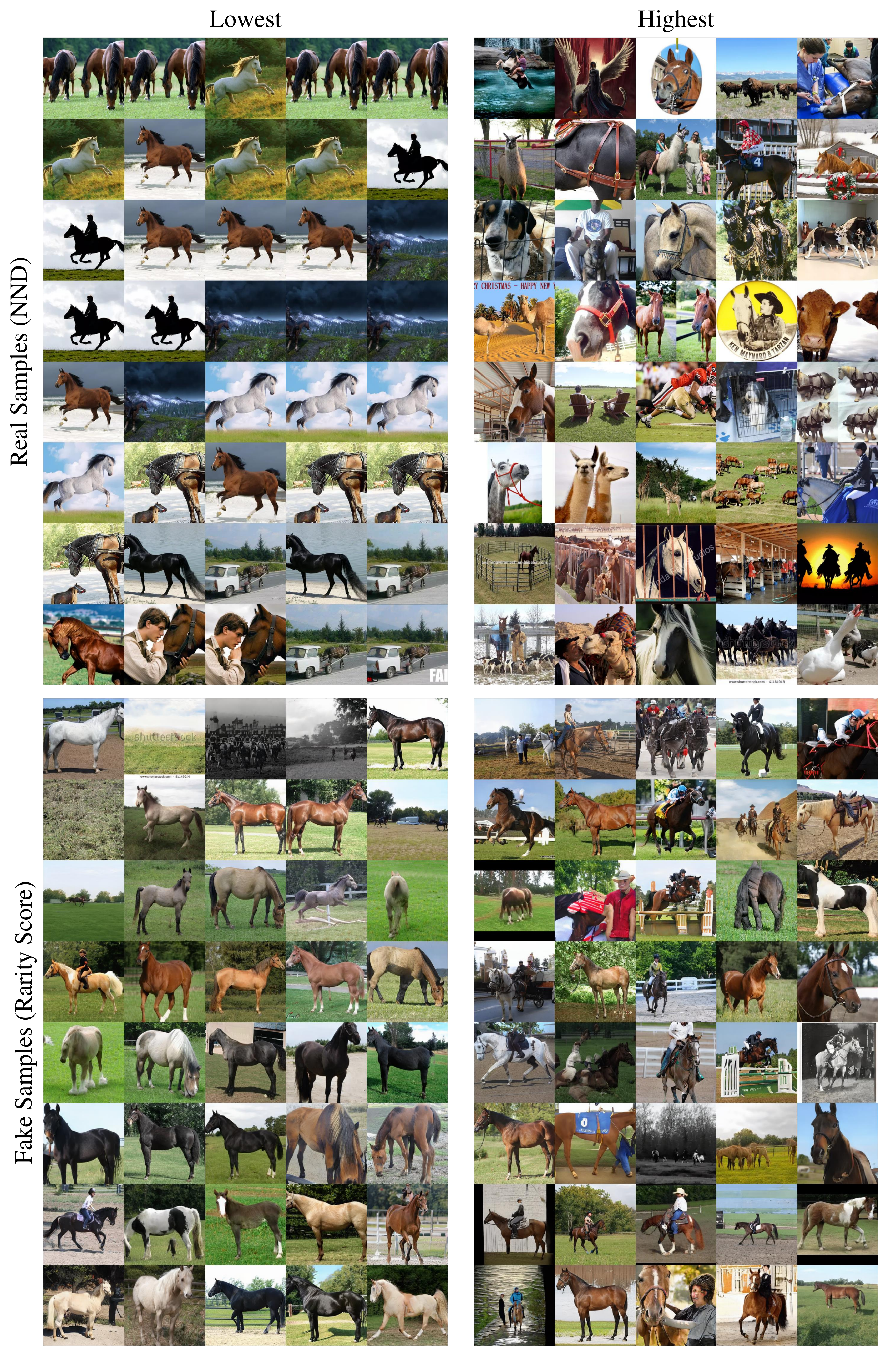}
         \caption{Real samples with the lowest/highest NND from LSUN-Horse dataset (top) and fake samples generated from StyleGAN2 with the lowest/highest rarity score (bottom).
         }
         \label{fig_apdx:qual_horse}
  
\end{figure}
\begin{figure}[H]
  \centering
  
         \includegraphics[width=\textwidth]{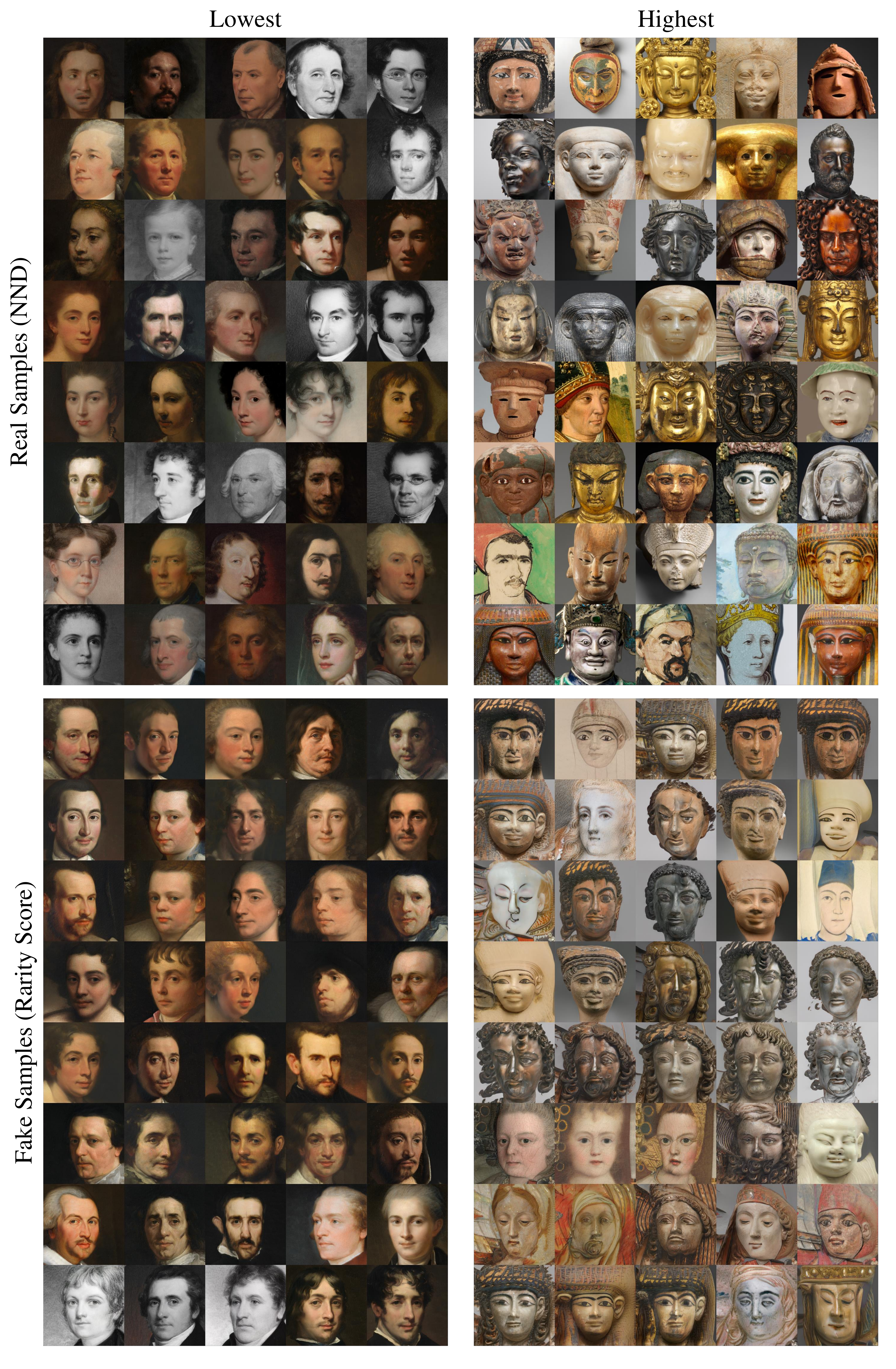}
         \caption{Real samples with the lowest/highest NND from MetFaces dataset (top) and fake samples generated from StyleGAN2-ADA with the lowest/highest rarity score (bottom).
         }
         \label{fig_apdx:qual_metface}
  
\end{figure}
\begin{figure}[H]
  \centering
  
         \includegraphics[width=\textwidth]{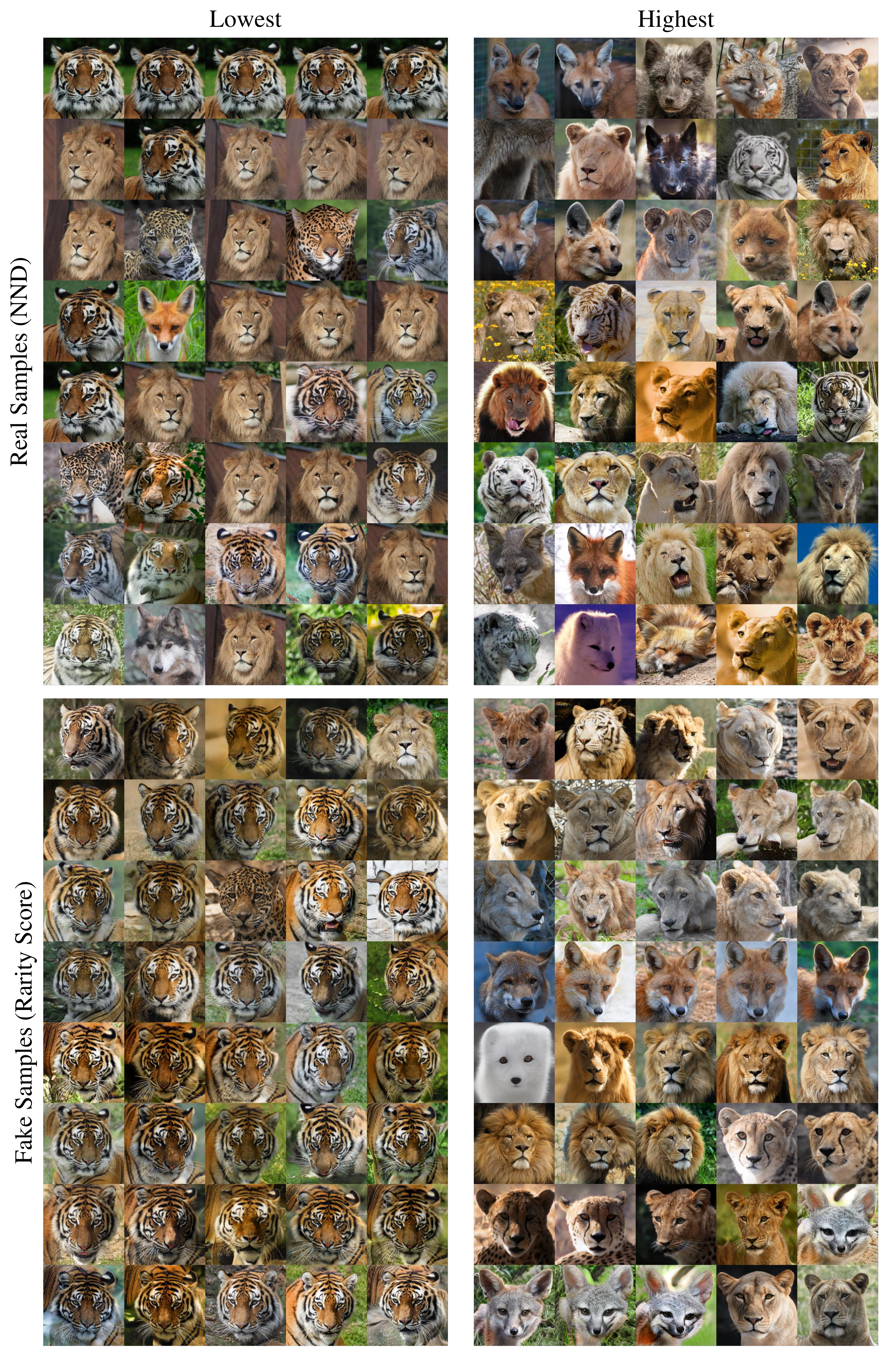}
         \caption{Real samples with the lowest/highest NND from AFHQ-Wild dataset (top) and fake samples generated from StyleGAN2-ADA with the lowest/highest rarity score (bottom).
         }
         \label{fig_apdx:qual_afhqwild}
  
\end{figure}

\newpage
\section{Examples of Out-of-Manifold Samples}
\label{app:oom}
\begin{figure}[!ht]
  \centering
  
         \includegraphics[width=\textwidth]{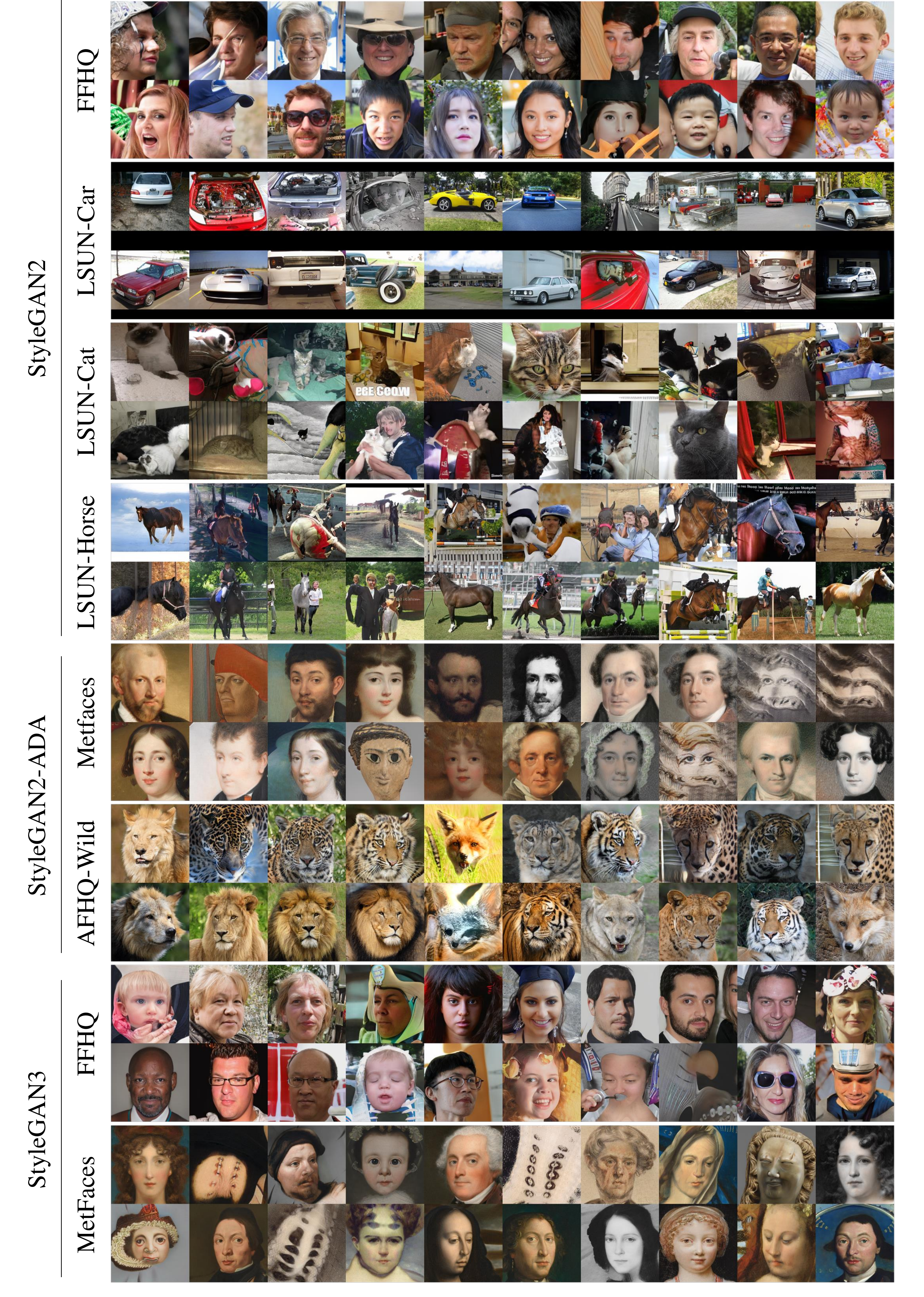}
         \caption{Out-of-manifold samples for various models.
         }
         \label{fig_apdx:oom}
  
\end{figure}

\newpage
\section{Results for the Various Feature Extractors}

\begin{figure}[ht!]
  \centering
  
         \includegraphics[width=0.9\textwidth]{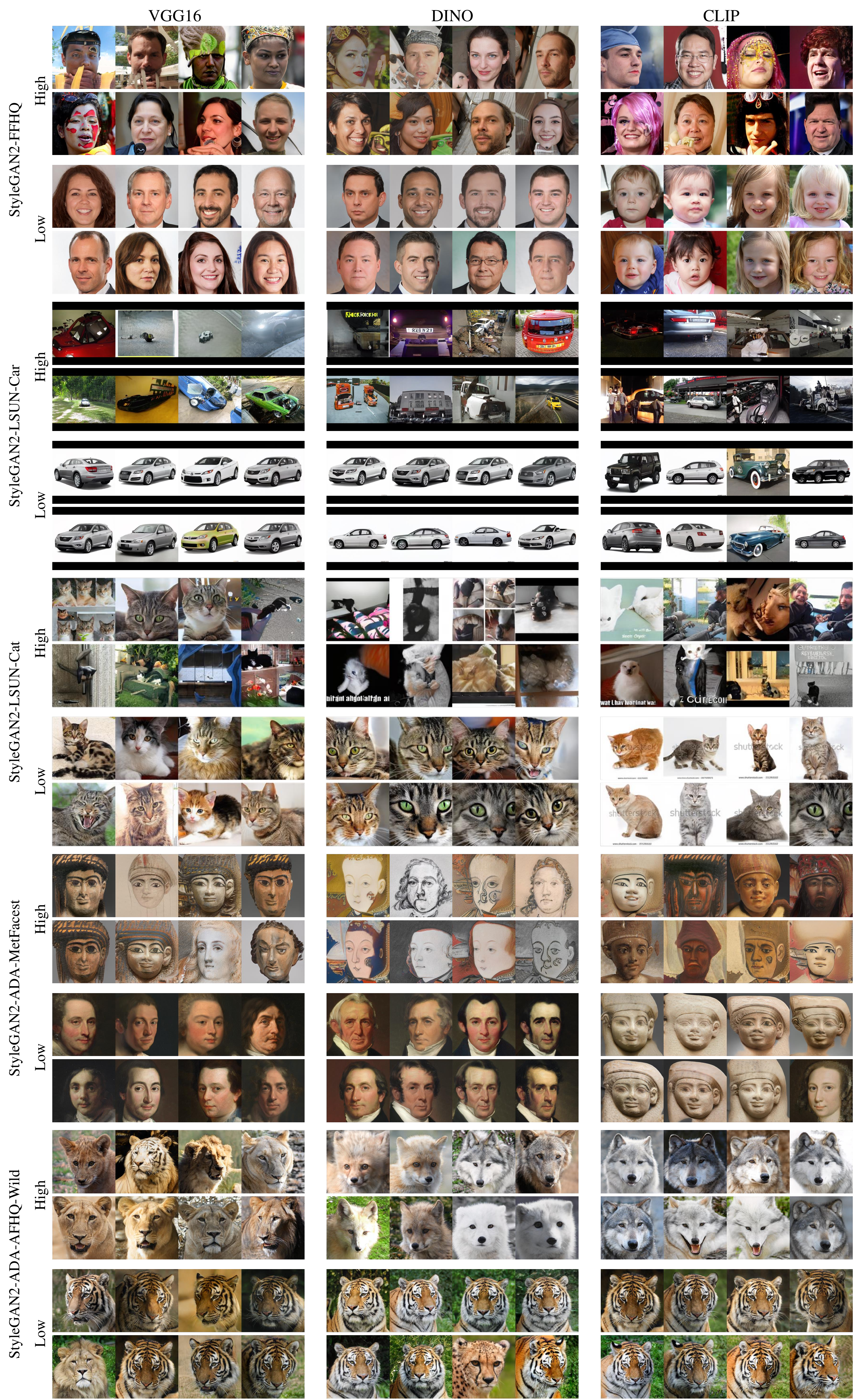}
         \caption{Fake samples with the lowest/highest rarity scores based on various feature extractors.
         }
         \label{fig_apdx:backbone}
  
\end{figure}

%% file: neurips_2022.bbl
\begin{thebibliography}{10}\itemsep=-1pt

\bibitem{simonyanZ14a}
Karen~Simonyan andAndrew Zisserman.
\newblock Very deep convolutional networks for large-scale image recognition.
\newblock In {\em ICLR}, 2015.

\bibitem{bau2018gan}
David Bau, Jun-Yan Zhu, Hendrik Strobelt, Bolei Zhou, Joshua~B Tenenbaum,
  William~T Freeman, and Antonio Torralba.
\newblock Gan dissection: Visualizing and understanding generative adversarial
  networks.
\newblock {\em ICLR}, 2019.

\bibitem{binkowski2018demystifying}
Mikołaj Bińkowski, Dougal~J. Sutherland, Michael Arbel, and Arthur Gretton.
\newblock Demystifying {MMD} {GAN}s.
\newblock In {\em ICLR}, 2018.

\bibitem{brock2018large}
Andrew Brock, Jeff Donahue, and Karen Simonyan.
\newblock Large scale {GAN} training for high fidelity natural image synthesis.
\newblock In {\em ICLR}, 2019.

\bibitem{caron2021emerging}
Mathilde Caron, Hugo Touvron, Ishan Misra, Herv\'e J\'egou, Julien Mairal,
  Piotr Bojanowski, and Armand Joulin.
\newblock Emerging properties in self-supervised vision transformers.
\newblock In {\em ICCV}, 2021.

\bibitem{choi2018stargan}
Yunjey Choi, Minje Choi, Munyoung Kim, Jung-Woo Ha, Sunghun Kim, and Jaegul
  Choo.
\newblock Stargan: Unified generative adversarial networks for multi-domain
  image-to-image translation.
\newblock In {\em CVPR}, 2018.

\bibitem{choi2020stargan}
Yunjey Choi, Youngjung Uh, Jaejun Yoo, and Jung-Woo Ha.
\newblock Stargan v2: Diverse image synthesis for multiple domains.
\newblock In {\em CVPR}, 2020.

\bibitem{Goodfellow2014generative}
Ian~J Goodfellow, Jean Pouget-Abadie, Mehdi Mirza, Bing Xu, David Warde-Farley,
  Sherjil Ozair, Aaron Courville, and Yoshua Bengio.
\newblock Generative adversarial networks.
\newblock In {\em NeurIPS}, 2014.

\bibitem{harkonen2020ganspace}
Erik H{\"a}rk{\"o}nen, Aaron Hertzmann, Jaakko Lehtinen, and Sylvain Paris.
\newblock Ganspace: Discovering interpretable gan controls.
\newblock {\em NeurIPS}, 2020.

\bibitem{heusel2017gans}
Martin Heusel, Hubert Ramsauer, Thomas Unterthiner, Bernhard Nessler, and Sepp
  Hochreiter.
\newblock Gans trained by a two time-scale update rule converge to a local nash
  equilibrium.
\newblock In {\em NeurIPS}, 2017.

\bibitem{isola2017image}
Phillip Isola, Jun-Yan Zhu, Tinghui Zhou, and Alexei~A Efros.
\newblock Image-to-image translation with conditional adversarial networks.
\newblock In {\em CVPR}, 2017.

\bibitem{jahanian2019steerability}
Ali Jahanian, Lucy Chai, and Phillip Isola.
\newblock On the ``steerability" of generative adversarial networks.
\newblock In {\em ICLR}, 2020.

\bibitem{karras2020training}
Tero Karras, Miika Aittala, Janne Hellsten, Samuli Laine, Jaakko Lehtinen, and
  Timo Aila.
\newblock Training generative adversarial networks with limited data.
\newblock In {\em NeurIPS}, 2020.

\bibitem{karras2021alias}
Tero Karras, Miika Aittala, Samuli Laine, Erik H{\"a}rk{\"o}nen, Janne
  Hellsten, Jaakko Lehtinen, and Timo Aila.
\newblock Alias-free generative adversarial networks.
\newblock {\em NeurIPS}, 2021.

\bibitem{karras2019style}
Tero Karras, Samuli Laine, and Timo Aila.
\newblock A style-based generator architecture for generative adversarial
  networks.
\newblock In {\em CVPR}, 2019.

\bibitem{karras2020analyzing}
Tero Karras, Samuli Laine, Miika Aittala, Janne Hellsten, Jaakko Lehtinen, and
  Timo Aila.
\newblock Analyzing and improving the image quality of stylegan.
\newblock In {\em CVPR}, 2020.

\bibitem{kim2021exploiting}
Hyunsu Kim, Yunjey Choi, Junho Kim, Sungjoo Yoo, and Youngjung Uh.
\newblock Exploiting spatial dimensions of latent in gan for real-time image
  editing.
\newblock In {\em CVPR}, 2021.

\bibitem{kim2019tag2pix}
Hyunsu Kim, Ho~Young Jhoo, Eunhyeok Park, and Sungjoo Yoo.
\newblock Tag2pix: Line art colorization using text tag with secat and changing
  loss.
\newblock In {\em ICCV}, 2019.

\bibitem{kim2021feature}
Junho Kim, Yunjey Choi, and Youngjung Uh.
\newblock Feature statistics mixing regularization for generative adversarial
  networks.
\newblock In {\em CVPR}, 2022.

\bibitem{Kim2020U-GAT-IT}
Junho Kim, Minjae Kim, Hyeonwoo Kang, and Kwang~Hee Lee.
\newblock U-gat-it: Unsupervised generative attentional networks with adaptive
  layer-instance normalization for image-to-image translation.
\newblock In {\em ICLR}, 2020.

\bibitem{kim2022c3gan}
Yunji Kim and Jung-Woo Ha.
\newblock Contrastive fine-grained class clustering via generative adversarial
  networks.
\newblock In {\em ICLR}, 2022.

\bibitem{kingma2018glow}
Durk~P Kingma and Prafulla Dhariwal.
\newblock Glow: Generative flow with invertible 1x1 convolutions.
\newblock In {\em NeurIPS}, 2018.

\bibitem{kynkaanniemi2019improved}
Tuomas Kynk{\"a}{\"a}nniemi, Tero Karras, Samuli Laine, Jaakko Lehtinen, and
  Timo Aila.
\newblock Improved precision and recall metric for assessing generative models.
\newblock In {\em NeurIPS}, 2019.

\bibitem{naeem2020reliable}
Muhammad~Ferjad Naeem, Seong~Joon Oh, Youngjung Uh, Yunjey Choi, and Jaejun
  Yoo.
\newblock Reliable fidelity and diversity metrics for generative models.
\newblock In {\em ICML}, 2020.

\bibitem{park2022how}
Namuk Park and Songkuk Kim.
\newblock How do vision transformers work?
\newblock In {\em ICLR}, 2022.

\bibitem{radford2021learning}
Alec Radford, Jong~Wook Kim, Chris Hallacy, Aditya Ramesh, Gabriel Goh,
  Sandhini Agarwal, Girish Sastry, Amanda Askell, Pamela Mishkin, Jack Clark,
  et~al.
\newblock Learning transferable visual models from natural language
  supervision.
\newblock In {\em ICML}, 2021.

\bibitem{salimans2016improved}
Tim Salimans, Ian Goodfellow, Wojciech Zaremba, Vicki Cheung, Alec Radford, and
  Xi Chen.
\newblock Improved techniques for training gans.
\newblock In {\em NeurIPS}, 2016.

\bibitem{shen2020interfacegan}
Yujun Shen, Jinjin Gu, Xiaoou Tang, and Bolei Zhou.
\newblock Interpreting the latent space of gans for semantic face editing.
\newblock In {\em CVPR}, 2020.

\bibitem{skorokhodov2021stylegan}
Ivan Skorokhodov, Sergey Tulyakov, and Mohamed Elhoseiny.
\newblock Stylegan-v: A continuous video generator with the price, image
  quality and perks of stylegan2.
\newblock 2022.

\bibitem{tian2021good}
Yu Tian, Jian Ren, Menglei Chai, Kyle Olszewski, Xi Peng, Dimitris~N Metaxas,
  and Sergey Tulyakov.
\newblock A good image generator is what you need for high-resolution video
  synthesis.
\newblock 2021.

\bibitem{tulyakov2018mocogan}
Sergey Tulyakov, Ming-Yu Liu, Xiaodong Yang, and Jan Kautz.
\newblock Mocogan: Decomposing motion and content for video generation.
\newblock In {\em CVPR}, 2018.

\bibitem{yu2022generating}
Sihyun Yu, Jihoon Tack, Sangwoo Mo, Hyunsu Kim, Junho Kim, Jung-Woo Ha, and
  Jinwoo Shin.
\newblock Generating videos with dynamics-aware implicit generative adversarial
  networks.
\newblock 2022.

\bibitem{Zhang_2018_CVPR}
Richard Zhang, Phillip Isola, Alexei~A. Efros, Eli Shechtman, and Oliver Wang.
\newblock The unreasonable effectiveness of deep features as a perceptual
  metric.
\newblock In {\em CVPR}, 2018.

\bibitem{zhu2017cyclegan}
Jun-Yan Zhu, Taesung Park, Phillip Isola, and Alexei~A Efros.
\newblock Unpaired image-to-image translation using cycle-consistent
  adversarial networks.
\newblock In {\em ICCV}, 2017.

\end{thebibliography}
